\documentclass[acmsmall,screen]{acmart}
\usepackage{graphicx} 
\usepackage{booktabs}
\usepackage{multirow}
\usepackage{hyperref}
\usepackage{float}
\usepackage{subcaption}
\usepackage{adjustbox}
\usepackage{tcolorbox}
\usepackage{fontawesome5}
\usepackage[utf8]{inputenc}
\usepackage[table]{xcolor}

\usepackage{tikz}
\usetikzlibrary{arrows.meta, positioning, calc, fit}

\renewcommand*{\equationautorefname}{Equation}
\def\equationautorefname~#1\null{(#1)\null}

\usepackage[vlined, boxruled, linesnumbered] {algorithm2e}
\SetKwComment{Comment}{$\triangleright$\ }{}
\SetCommentSty{itshape}
\newboolean{showcomments}
\setboolean{showcomments}{true}
\ifthenelse{\boolean{showcomments}}
{
	\definecolor{myyellow}{RGB}{255, 228, 26}
	\definecolor{myblue}{RGB}{50, 50, 220}
	
}

\newcommand{\tool}{\textsc{MoST}\xspace}

\title{Search-Based Testing of Modality Sensitivity in Vision Language Models}

\author{Oliver Wei{\ss}l}
\orcid{0009-0008-7575-0187}
\email{o.weissl@tum.de}
\affiliation{%
  \institution{Technical University of Munich}
  \city{Garching near Munich}
  \country{Germany}
}

\author{Eduardo Breitembach}
\orcid{0009-0004-5722-4694}
\email{e.breitembach@tum.de}
\affiliation{%
  \institution{Technical University of Munich}
  \city{Garching near Munich}
  \country{Germany}
}

\author{Andrea Stocco}
\orcid{0000-0001-8956-3894}
\email{andrea.stocco@tum.de}
\affiliation{%
  \institution{Technical University of Munich}
  \city{Garching near Munich}
  \country{Germany}
}
\email{stocco@fortiss.org}
\affiliation{%
  \institution{fortiss GmbH}
  \city{Munich}
  \country{Germany}
}

\newcommand{\head}[1]{\noindent\textbf{#1.}}
\newtheorem*{definition}{\bf Definition}

\begin{document}

\begin{abstract}
Vision-language models (VLMs) process images and text jointly, yet
robustness evaluations often perturb these inputs separately or
assess only predefined corruption levels. Such evaluations provide
limited evidence of which modality is more susceptible to
perturbations and whether smaller changes across both modalities
can jointly expose failures.
We present MoST, a black-box testing framework that characterizes
modality sensitivity by comparing image-only, text-only, and joint
perturbations. MoST uses multi-objective search to select and combine
perturbations and adjust their severity, seeking prediction
degradation while limiting changes to images and prompts. 
We evaluate MoST on four VLMs performing object localization on
ImageNet and driving images. The models exhibit distinct
sensitivities: textual perturbations dominate for some, whereas
others are susceptible across both modalities. In several
configurations, joint perturbations match or exceed the effectiveness
of single-modality perturbations with smaller modifications to each
input. In a human study, 60--76\% of assessments across models
judge the perturbed inputs sufficiently understandable to perform
the task, and human localization is more accurate than that of the
corresponding VLMs. These results show that evaluating modalities
separately can overlook failure-inducing combinations and that
modality sensitivity provides a more detailed account of VLM
robustness than aggregate failure rates alone.
\end{abstract}


\maketitle

\section{Introduction}\label{sec:intro}

Vision-Language Models (VLMs) combine visual observations and natural
language to perform tasks such as object localization, visual question
answering, and scene understanding~\cite{zhang2024vision}. Their
application to autonomous driving and other safety-critical settings
requires systematic evaluation of their behavior under degraded
inputs~\cite{10531702,orjuela2026improvingrobustnessvisionlanguageactionmodels}.
Prediction quality depends on both the information available in each
modality and the model's ability to interpret the two inputs jointly.

Visual and textual inputs can deteriorate independently or
simultaneously. Images may be affected by adverse weather, sensor
noise, motion blur, or compression
artifacts~\cite{lambertenghi2025benchmarkingimageperturbationstesting}.
Textual instructions may contain transcription errors introduced by
speech recognition~\cite{AHLAWAT2025201}, as well as spelling errors,
ambiguity, or changes in meaning introduced by users or upstream
components. Evaluating these conditions requires distinguishing
susceptibility to visual perturbations, textual perturbations, and
their combination. This distinction can help developers identify
which input conditions require additional testing and mitigation.

Existing robustness evaluations commonly measure performance under
predefined corruption levels or assess perturbations to individual
modalities~\cite{hendrycks2019robustness,
sui2025benchmarkingcorruptionrobustnesslvlms,
qiu2024benchmarkingrobustnessmultimodalimagetext}.
Although these evaluations quantify performance degradation, aggregate
scores can conceal differences in the perturbations required to expose
failures. Two models with similar failure rates may differ substantially
in their susceptibility to image and text modifications. Moreover,
perturbing both modalities may expose failures with smaller changes
to each input than perturbing either modality alone.

We refer to this behavior as \emph{modality sensitivity}: the relative
susceptibility of a VLM to visual, textual, and joint input degradation,
considering both prediction degradation and the extent of the applied
perturbations. Characterizing modality sensitivity requires comparisons
across these three settings and it also requires exploring perturbation
combinations and severities beyond a small set of predefined levels,
since the modifications needed to expose a failure may differ across
models and inputs.

To this aim, we propose \tool (\textbf{Mo}dality-\textbf{S}ensitivity
\textbf{T}esting), a search-based black-box testing framework for
characterizing modality sensitivity in VLMs. Starting from an image and its
task prompt, \tool searches for combinations of perturbations and
adjusts their severity to degrade the VLM predictions while limiting
visual and textual changes. The search is performed separately under
image-only, text-only, and joint perturbation regimes. Comparing the
resulting test cases produces a behavioral profile of the model's
susceptibility under the evaluated perturbations, without requiring
access to its parameters or gradients.


We evaluate \tool on four recent VLMs---Qwen, Kimi, InternVL, and
Nemotron---performing object localization on ImageNet and driving
images. The results reveal distinct modality-sensitivity profiles.
Qwen exhibits greater susceptibility to textual than visual
perturbations, whereas Kimi exhibits failures across all three
regimes, frequently involving malformed outputs. InternVL and
Nemotron show patterns that vary with the scene type and failure
criterion. In several configurations, joint perturbations achieve
failure rates comparable to or higher than those of single-modality
perturbations while requiring smaller modifications to each input.
A human study further shows that most assessments find the perturbed inputs interpretable, with humans achieving higher localization accuracy than the corresponding VLMs.


\noindent
This paper makes the following contributions:
\begin{enumerate}

    \item \textbf{Testing framework.}
    \tool, a search-based black-box framework that characterizes
    modality sensitivity through image-only, text-only, and joint
    perturbations, optimizing prediction degradation alongside
    visual and textual fidelity~\cite{replication-package}.

    \item \textbf{Comparative measure.}
    Stealth-Weighted Task Degradation (SWTD), a novel metric combining relative task degradation and
    perturbation magnitude to support comparisons across
    single-modality and joint perturbation regimes.

    \item \textbf{Empirical evaluation.}
    A study of four VLMs across ImageNet and driving scenes,
    examining failure rates, localization degradation, perturbation
    requirements, and computational cost, complemented by a human
    assessment of test-case interpretability.

\end{enumerate}

\section{Background}\label{sec:background}

\subsection{Modality Sensitivity in Vision-Language Models}

VLMs jointly process visual and textual inputs to perform
downstream tasks such as visual question answering, object localization, and scene
understanding~\cite{zhang2024vision}. Unlike uni-modal models, prediction quality
depends not only on the information contained in each modality, but also on how the
model combines these complementary sources through its cross-modal fusion
mechanism. 
Degradation in either or both modalities can affect predictions
differently, motivating separate and joint evaluation.

\begin{definition}[\textbf{Modality Sensitivity}]
Modality sensitivity characterizes a vision-language model's
susceptibility to visual, textual, and joint input degradation.
Given an input pair $(I,T)$ and perturbed variants $(I',T)$,
$(I,T')$, and $(I',T')$, it is assessed through the relative
prediction degradation and perturbation effort required under
image-only, text-only, and joint perturbations.
\end{definition}

This behavioral perspective naturally distinguishes three complementary cases:

\begin{itemize}
    \item \textbf{Visual sensitivity:} failures are induced more readily through visual perturbations;
    \item \textbf{Textual sensitivity:} failures are induced more readily through textual perturbations;
    \item \textbf{Cross-modal sensitivity:} joint perturbations expose failures that are not equivalently revealed by either uni-modal regime alone.
\end{itemize}

\subsection{Modality Sensitivity Testing}

Traditional robustness evaluation measures whether prediction quality degrades
under perturbed inputs. While such evaluations quantify overall robustness, they
provide limited insight into which modality primarily contributes to the observed
degradation. Consequently, two VLMs may exhibit similar aggregate robustness while
having fundamentally different modality-sensitivity profiles.

We refer to the complementary objective of characterizing this behavior as
\emph{modality sensitivity testing}. Rather than producing a single robustness
score, it compares failure rates, degradation severity, and perturbation
requirements across visual, textual, and joint perturbation regimes, providing
developers with information about model-specific weaknesses that cannot be
obtained from uni-modal robustness measurements alone.

Since image perturbations, text perturbations, and their continuous severity
combinations define a large joint search space, exhaustive evaluation quickly
becomes infeasible. This motivates the search-based methodology presented in the
next section.
\section{Methodology}\label{sec:method}

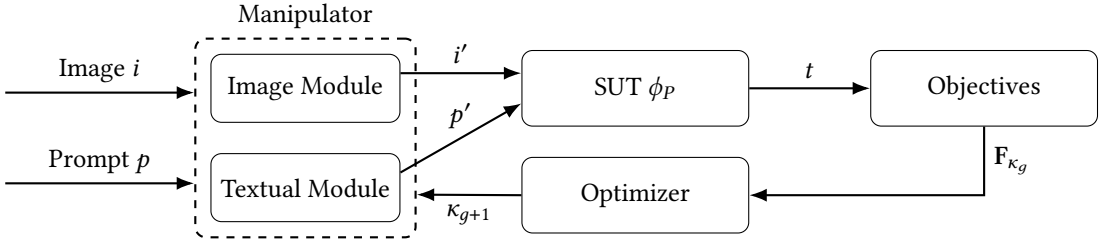
\begin{figure*}[t]
\centering
\begin{tikzpicture}[
    block/.style={rectangle, draw, rounded corners, minimum width=3.0cm, minimum height=1.0cm, align=center},
    smallblock/.style={rectangle, draw, rounded corners, minimum width=2.5cm, minimum height=0.9cm, align=center},
    arrow/.style={-Latex, thick},
    node distance=1.6cm and 1.6cm
]

\node[smallblock] (imgmod) {Image Module};
\node[smallblock, below=0.4cm of imgmod] (txtmod) {Textual Module};

\node[draw, dashed, thick, rounded corners, fit=(imgmod)(txtmod),
      label={Manipulator}, inner sep=0.2cm] (manip) {};

\node[block, right=of imgmod] (sut) {SUT $\phi_P$};
\node[block, right=of sut] (obj) {Objectives};
\node[block, below=0.4cm of sut] (opt) {Optimizer};

\draw[arrow] ($(manip.west)+(-2.5,0.6)$) -- ($(manip.west)+(0.0,0.6)$)
    node[pos=0.5, above] {Image $i$};

\draw[arrow] ($(manip.west)+(-2.5,-0.6)$) -- ($(manip.west)+(0.0,-0.6)$)
    node[pos=0.5, above] {Prompt $p$};

\draw[arrow] ($(imgmod.east)+(0.0,0.2)$) -- ($(sut.west)+(0.0,0.2)$)
    node[pos=0.5, above] {$i'$};
\draw[arrow] ($(txtmod.east)+(0.0,0.2)$) -- ($(sut.west)+(0.0,-0.2)$)
    node[pos=0.5, above] {$p'$};

\draw[arrow] (sut.east) -- (obj.west)
    node[pos=0.5, above] {$t$};

\draw[arrow] (obj.south) |- (opt.east)
    node[pos=0.25, right] {$\mathbf{F}_{\kappa_{g}}$};

\draw[arrow] (opt.west) -- ($(manip.east)+(0.0,-0.75)$)
    node[pos=0.5, below] {$\kappa_{g+1}$};

\end{tikzpicture}
\caption{System overview. The Manipulator (\autoref{sec:manip}) perturbs both image and prompt inputs. The SUT (\autoref{sec:sut}) processes the modified inputs, whose outputs are evaluated by the objective functions (\autoref{sec:obj}). The optimizer (\autoref{sec:optim}) updates the manipulation strategy $\kappa$, controlling the manipulation process, based on the objective values.}
\label{fig:manipulator_loop}
\end{figure*}

The objective of \tool is to characterize the \emph{modality sensitivity} of vision-language models by systematically generating failure-inducing test cases under controlled input degradation. To this end, \tool explores three perturbation regimes: (i) image-only perturbations, (ii) text-only perturbations, and (iii) joint image-text perturbations. Comparing the failures discovered under these regimes characterizes the model's susceptibility to visual, textual, and joint perturbations.

Figure~\ref{fig:manipulator_loop} presents the overall workflow. Starting from a seed image $i$ and prompt $p$, the manipulator generates perturbed inputs, the system under test produces predictions, the objectives evaluate prediction degradation and perturbation fidelity, and the optimizer searches for failure-inducing test cases.

\subsection{System Under Test}\label{sec:sut}

The system under test (SUT) is a VLM that jointly processes an
image $i$ and a task prompt $p$ to produce a textual output $t$,
formally $\phi_P : \mathcal{I} \times \mathcal{P} \rightarrow
\mathcal{T}$. In our evaluation, the prompt requests structured
object-localization results. The initial prompt follows a fixed
template populated with the requested class labels. \tool perturbs
the image, the prompt, or both to evaluate modality sensitivity.

\begin{table}[t]
\caption{Visual manipulation methods.}
\label{tab:visual_manipulations}
\centering

\begin{tabular}{ll}
\toprule
\footnotesize
\textbf{Group} & \textbf{Manipulation Types} \\
\midrule
Digital &
JPEG Compression, Pixelation \\
Blur &
Defocus Blur, Motion Blur \\
Noise &
Gaussian Noise \\
Color &
HSV Hue Shift, Grayscale Blending, Contrast Manipulation \\
Geometric &
Elastic Transformation, Bounding-Box-Aware Cutouts \\
Weather / Environmental &
Fog Overlay, Snow Overlay \\
\bottomrule
\end{tabular}

\end{table}

\begin{figure*}[t]
    \centering
    \begin{tabular}{@{}c@{}c@{}c@{}c@{}c@{}c@{}}
    \includegraphics[width=0.15\linewidth, trim=411.04pt 50pt 411.04pt 0, clip]{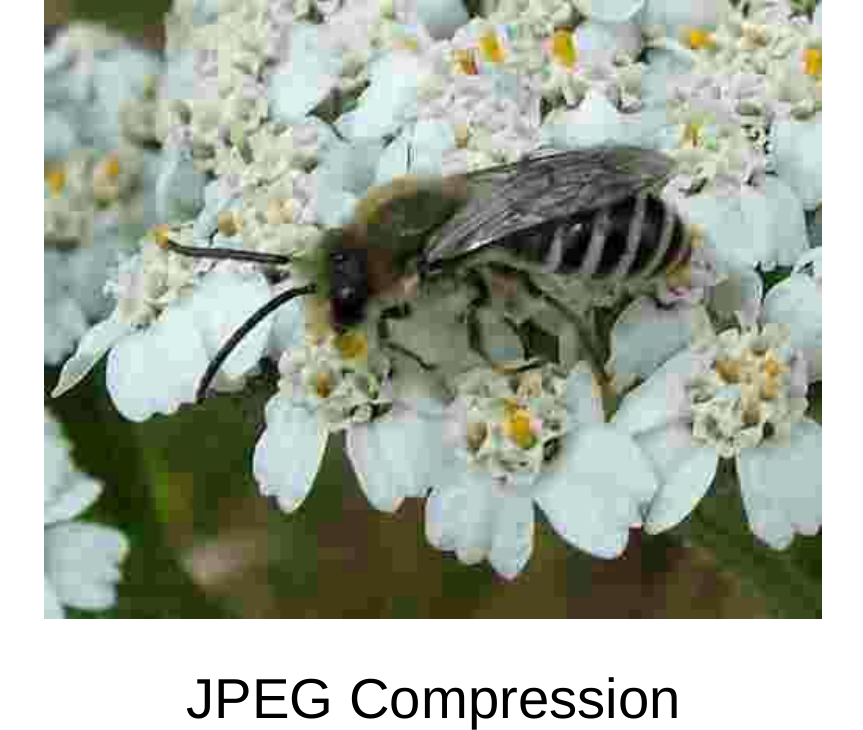} &
    \includegraphics[width=0.15\linewidth, trim=549.28pt 34pt 0 0, clip]{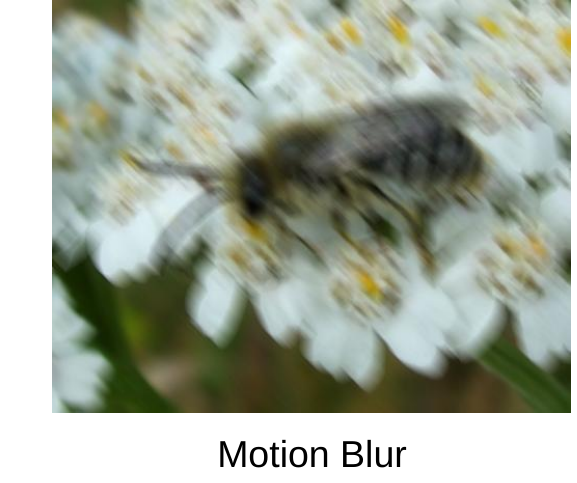} &
    \includegraphics[width=0.15\linewidth, trim=374.5pt 45.55pt 0 0, clip]{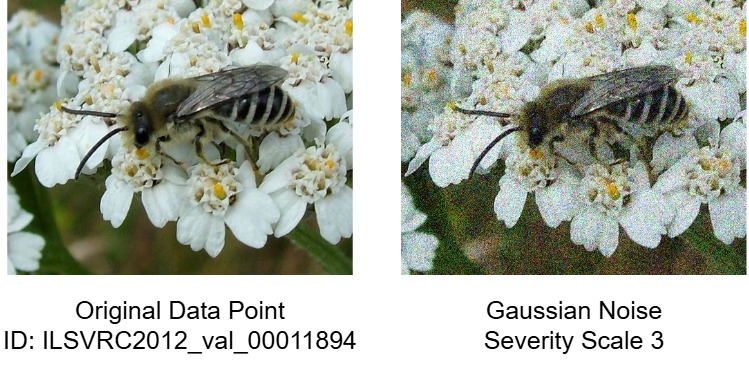} &
    \includegraphics[width=0.15\linewidth, trim=1153.5pt 69pt 576.75pt 0, clip]{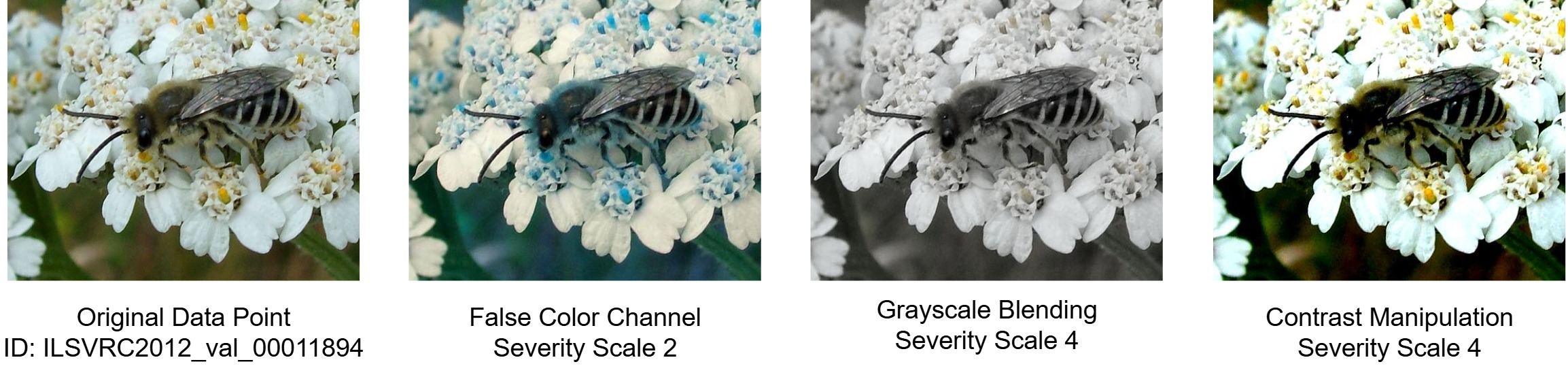} &
    \includegraphics[width=0.15\linewidth, trim=1145.33pt 69pt 0 0, clip]{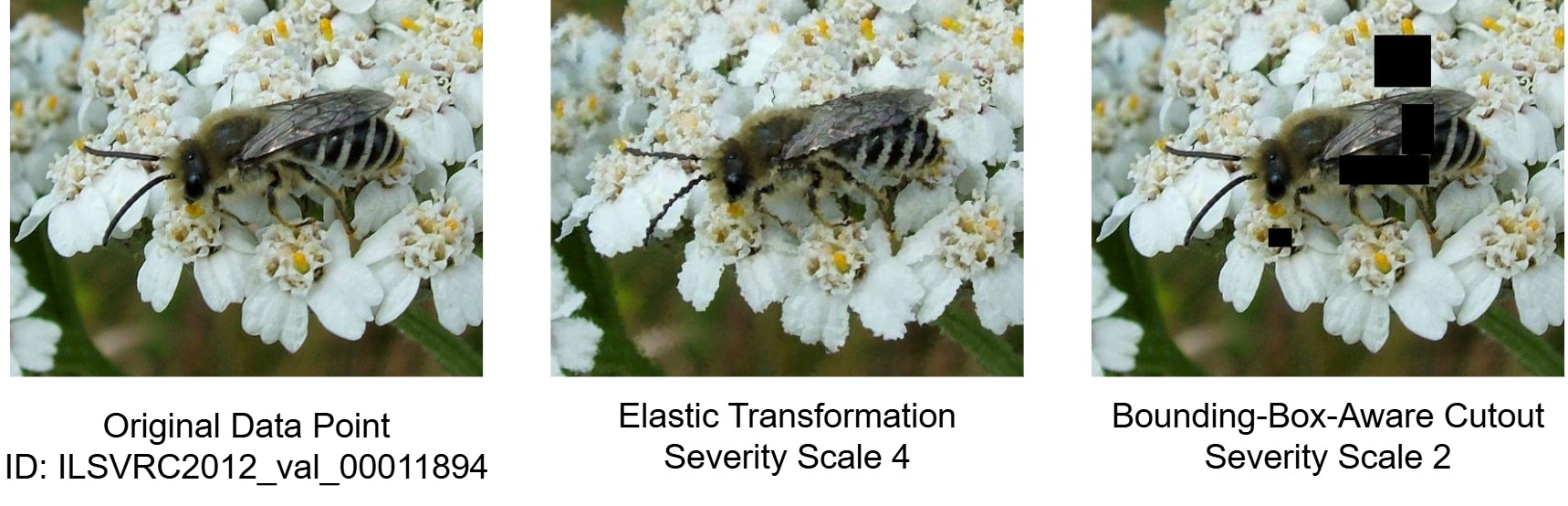} &
    \includegraphics[width=0.15\linewidth, trim=1143.33pt 69pt 10pt 0, clip]{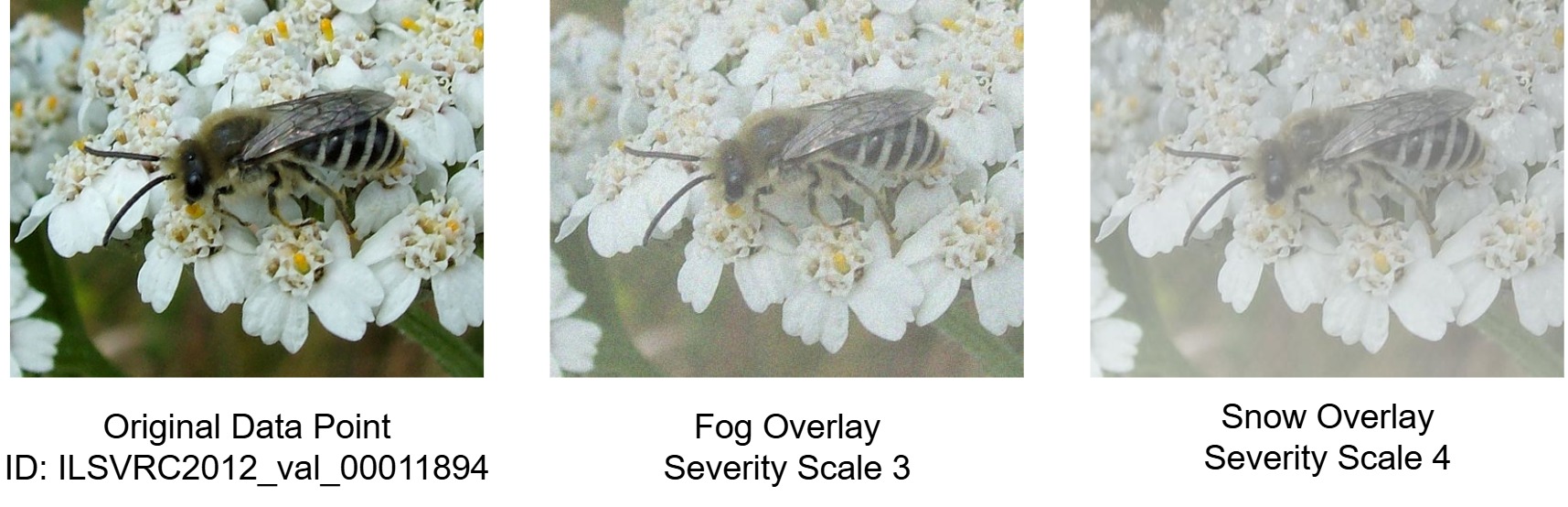} \\
    \end{tabular}
    \caption{Examples of each visual manipulation group from \autoref{tab:visual_manipulations}.}
    \label{fig:manip_gallery}
\end{figure*}

\subsection{Manipulator}\label{sec:manip}

The manipulator is responsible for generating controlled perturbations that expose the contribution of each modality to model failures. Depending on the selected perturbation regime, perturbations are applied to the image only, the text only, or both modalities simultaneously.

More in detail, it receives as input an image $i$, a textual prompt $p$, and a strategy vector $\kappa$ that parameterizes the applied transformations. The strategy $\kappa$ is initialized close to zero to allow gradual adaptation during optimization. The manipulator produces a perturbed image $i'$ and a perturbed prompt $p'$, where each perturbation is optional and applied only if specified by $\kappa$. This design ensures that the manipulator represents both identity mappings (no perturbation) and progressively stronger modifications within the same framework.

\subsubsection{Image-based component}

The image module applies visual transformations from \autoref{tab:visual_manipulations}, proposed by Lambertenghi et al.~\cite{lambertenghi2025benchmarkingimageperturbationstesting} to cover commonly used input degradations: compression artifacts, blur, noise, hue shifts, geometric distortions, and environmental overlays (e.g., defocus/motion blur via convolutional kernels, elastic warping that preserves semantic content, cutout occlusions, and physically motivated fog/snow; \autoref{fig:manip_gallery} shows representative examples). The strategy $\kappa$ determines which transformations are applied, at what intensity, and in what combination. Each transformation is controlled by a continuous severity parameter $S \in [0,1]$, modulating transformation-specific properties (e.g., blur radius or noise variance) via interpolation between the predefined levels of Lambertenghi et al.~\cite{lambertenghi2025benchmarkingimageperturbationstesting}.


\subsubsection{Textual component}

The textual module perturbs the prompt (\autoref{tab:textual_manip}) along three axes: orthographic (character swaps, e.g., Latin-to-Cyrillic; zero-width insertions; visually similar substitutions, affecting tokenization while preserving readability), semantic (synonym or phonetically similar replacements that preserve linguistic plausibility), and instruction-level (task-irrelevant prefixes or suffixes around the unchanged task instruction). Application and strength are governed by $\kappa$; the severity parameter $S \in [0,1]$ maps to a perturbation probability $P = 0.1 + 0.8S$ governing how frequently and strongly tokens are modified.


\begin{table}[tb]
\centering
\caption{Overview of Textual Manipulation Methods.}

\begin{tabular}{lll}
\toprule\footnotesize
\textbf{Group} & \textbf{Type} & \textbf{Example} \\
\midrule
\multirow{3}{*}{Orthographic}
 & Fragmentation & \texttt{crutch $\rightarrow$ cru tch} \\
 & Character Noise &\texttt{guitar $\rightarrow$ gu\_i\textbf{\scriptsize\textbackslash u200d}ta\textbf{\scriptsize\textbackslash u200b}r} \\
 & Saliency & \texttt{pepper $\rightarrow$ pebper} \\
\midrule
\multirow{2}{*}{Semantic}
 & Synonym & \texttt{elephant $\rightarrow$ pachyderm} \\
 & Homophone & \texttt{elephant $\rightarrow$ ella font} \\
\midrule
\multirow{3}{*}{Instruction-Level}
 & Adversarial Suffix & \texttt{\textbf{[Task]} + "!@\# ignore"} \\
 & Context Rot & \texttt{garbled prefix + \textbf{[Task]}} \\
 & Task Reinforcement & \texttt{"verify carefully" + \textbf{[Task]}} \\
\bottomrule
\end{tabular}
\label{tab:textual_manip}
\end{table}

\subsection{Objectives}\label{sec:obj}

Characterizing modality sensitivity requires failures that remain meaningful and realistic. Otherwise, failures may simply result from excessive corruption rather than dependence on a particular modality. Consequently, our search optimizes three competing objectives: maximizing prediction degradation while simultaneously preserving visual realism and semantic consistency.
Together, these three objectives characterize the effectiveness and severity of a manipulation. As these goals are inherently conflicting, the optimization is formulated as a multi-objective minimization problem.

Given an input image $i$, a prompt $p$, and perturbed inputs $i'$ and $p'$, the objective vector is defined as:
\begin{equation}
    \mathbf{F}_\kappa =
    \begin{pmatrix}
        f_1(\mathcal{B}, \mathcal{G}) \\
        f_2(i, i') \\
        f_3(p, p')
    \end{pmatrix} =
    \begin{pmatrix}
        \autoref{eq:iou} \\
        \autoref{eq:frobenius} \\
        \autoref{eq:text-sim}
    \end{pmatrix}
\end{equation}
where $\mathcal{B}$ denotes the predicted bounding boxes of the SUT from input $i', p'$ and $\mathcal{G}$ the ground truth annotations.

\head{Detection degradation} It is measured as the Intersection over Union (IoU) between the predicted bounding boxes $\mathcal{B}$ and the ground truth boxes $\mathcal{G}$ of an input, under a one-to-one assignment. We compute the IoU for every pair of predicted and ground truth boxes and determine the assignment $\pi^{*}$ that maximizes the total IoU using the Hungarian method~\cite{kuhn1955hungarian}, so that each predicted box is matched to at most one ground truth box and vice versa:
\begin{equation}\label{eq:iou}
  f_1(\mathcal{B}, \mathcal{G}) = \frac{1}{\max(|\mathcal{B}|, |\mathcal{G}|)} \sum_{(j,k) \in \pi^{*}} \mathrm{IoU}\bigl(b_{j},\; g_{k}\bigr).
\end{equation}
Unmatched predicted or ground truth boxes contribute zero, so spurious, missing, or merged boxes all lower the score, and an empty output yields $f_1=0$. We deliberately ignore predicted labels: textual perturbations such as synonyms or homophones may legitimately change how the model names an object, and label matching would then count a correctly localized object as a failure. The metric therefore measures localization quality only.

\head{Visual fidelity} The second objective $f_2$ measures the deviation between the original and perturbed image using a normalized Frobenius distance \autoref{eq:frobenius}.
\begin{equation}\label{eq:frobenius}
    f_2(i, i') = \frac{\lVert i - i' \rVert_F}{\sqrt{H\cdot W\cdot C}},
\end{equation}
 
This requires the ranges of the image pairs to be the same, which in this work are $\in \left[0,1\right]$.
Minimizing this objective penalizes visual changes, but does not
guarantee perceptual similarity or preservation of scene content.

\head{Semantic fidelity} It constitutes the third objective $f_3$, which measures the semantic deviation between the original and perturbed prompt as the cosine distance of their text embeddings \autoref{eq:text-sim}, where $0$ indicates identical meaning, $1$ indicates no semantic overlap, and $\psi(\cdot)$ denotes an embedding model. In this work, we use Qwen3-Embedding-0.6B~\cite{qwen3embedding}.

\begin{equation}\label{eq:text-sim}
    f_3(p, p') = d_{\cos}(\psi(p),\; \psi(p')\bigr).
\end{equation}

Minimizing this objective encourages similarity between the prompt
embeddings, providing an approximate measure of semantic preservation.
Since no single solution can simultaneously optimize all objectives, the optimizer searches for a set of non-dominated solutions that represent trade-offs between degradation and fidelity.

\subsubsection{Early stopping}
To reduce computational cost, the optimization is terminated once a candidate solution satisfies all objective thresholds simultaneously. Concretely, the search stops when a strategy $\kappa$ fulfills:
\[
    (f_1 \leq \tau_{\mathrm{IoU}}) \land
    (f_2 \leq \tau_{\mathrm{img}}) \land
    (f_3 \leq \tau_{\mathrm{txt}})
\]
indicating sufficient degradation while maintaining acceptable visual and semantic fidelity.
The resulting Pareto-optimal solutions represent realistic failure-inducing test cases that characterize the model's modality sensitivity.

\subsection{Optimizer}\label{sec:optim}

The optimizer searches for perturbation strategies $\kappa$ that reveal the
relative contribution of the visual and textual modalities to model failures.
Rather than maximizing perturbation magnitude, it seeks minimal modifications that
induce failures while satisfying the fidelity objectives.

A strategy is a continuous vector $\kappa = [\kappa_{\mathrm{img}},
\kappa_{\mathrm{txt}}]$, split into an image- and a text-related component. Each
component contains one parameter in $[0,1]$ per manipulation type
(\autoref{tab:visual_manipulations}, \autoref{tab:textual_manip}), defining the
severity of that perturbation. All perturbations are applied within their
respective modality and composed into a single multi-modal transformation, so that
in contrast to selecting one perturbation at a time, the optimizer can explore
interaction effects between perturbation types. The resulting inputs are evaluated
by the objective functions (\autoref{sec:obj}).

To prevent trivial solutions in which all perturbations are maximized, the
optimizer enforces a total budget $B_{\mathrm{img}} + B_{\mathrm{txt}} \leq B$,
with $B=1$ in our experiments. In uni-modal configurations the entire budget is
available to the active modality; in multi-modal configurations the optimizer
distributes it across both.

The initial population is generated using budget-aware sampling: a total budget
usage is sampled from $[0,B]$ and distributed across the active modalities and
their perturbation parameters using a symmetric Dirichlet
distribution~\cite{fang2018symmetric}, yielding diversity in both total
perturbation strength and its allocation across manipulation types. During
optimization, candidate strategies are modified through evolutionary operators,
followed by a repair step that clamps all parameters to $[0,1]$ and, where a
modality exceeds its budget, rescales its values proportionally.

Executing the search independently under image-only, text-only, and joint regimes
produces complementary sets of failure-inducing test cases, whose comparison yields
a modality-sensitivity profile indicating whether failures arise from visual
perturbations, textual perturbations, or their interaction.

\section{Empirical Study}\label{sec:study}

\subsection{Research Questions}\label{sec:rqs}

\textbf{RQ\textsubscript{1} (effectiveness):}
How effective is \tool at generating failure-inducing test cases for VLMs?

We assess failure discovery under image-only, text-only, and joint
perturbation regimes, considering both prediction degradation and
the extent of visual and textual modifications.

\textbf{RQ\textsubscript{2} (efficiency):}
What computational effort does \tool require to generate
failure-inducing test cases?

We quantify search effort and runtime, together with the perturbation
budgets used by each regime.

\textbf{RQ\textsubscript{3} (validity):}
To what extent do generated test cases remain interpretable to humans?

We assess whether human evaluators can perform the requested
localization task on the perturbed inputs and compare their
localization accuracy with that of the corresponding VLMs.

\textbf{RQ\textsubscript{4} (modality sensitivity):}
How does susceptibility to visual, textual, and joint perturbations
vary across VLMs?

We characterize modality sensitivity by comparing attack success,
localization degradation, and perturbation requirements across
models and perturbation regimes.

\subsection{Objects of Study}

We evaluate recent VLMs on a bounding-box detection task using ImageNet-based inputs from the ILSVRC 2012 validation dataset~\cite{imagenet} and real-world Driving scenes from the Udacity dataset.\footnote{\url{https://github.com/udacity/self-driving-car/tree/master/datasets}} 
The models retained for the final analysis are Qwen3-VL-4B-Instruct~\cite{Qwen3-VL}, Kimi-VL-A3B-Instruct~\cite{kimiteam2025kimivltechnicalreport}, Nemotron3 Nano Omni-30B~\cite{nvidia2026nemotron3nanoomni}, and InternVL3.5-8B~\cite{wang2025internvl3_5}.

We additionally considered Gemma-3-4B-IT~\cite{gemma_2025} and DeepSeek-VL2-Tiny~\cite{wu2024deepseekvl2mixtureofexpertsvisionlanguagemodels} during preliminary experiments. Both exhibited very high baseline failure rates ($\geq 0.95$), leaving too few baseline-valid inputs for a meaningful modality-sensitivity analysis. Therefore, they were excluded from certain parts of the analysis due to lack of data.
All experiments were conducted under a single-GPU hardware constraint using one NVIDIA L40S GPU with 48 GB of VRAM. All evaluated models fit within the memory capacity of this device.

For all models, we use the prompt template:
\begin{center}
\vspace{-2pt}
\fbox{\parbox{0.95\linewidth}{\small\textit{Detect the object(s) ``[class labels]'' in this image. Return ONLY a JSON array --- no markdown, no commentary. Each element must have exactly two keys: ``label'' (string, one of the requested objects) and ``bbox'' ([x1, y1, x2, y2] bounding box coordinates). Example: [\{``label'': ``cat'', ``bbox'': [10, 20, 150, 200]\}, \{``label'': ``dog'', ``bbox'': [300, 50, 480, 400]\}]}}}
\vspace{-2pt}
\end{center}
Here, \textit{[class labels]} is replaced with the comma-separated ground-truth object classes associated with the input image.

\subsection{Metrics and Measurements}

\subsubsection{Effectiveness Metrics (RQ\textsubscript{1})}\label{sec:rq1_metrics}
Effectiveness is measured using the attack success rate at threshold $\tau$.
Let $\mathcal{X}$ denote the selected set of evaluated test cases and let
$\mathcal{F}_{\tau}(x)$ denote the chosen failure criterion for test case $x$.
We define

\begin{equation}
\mathrm{ASR}_{\tau}(\mathcal{X},\mathcal{F})
=
\frac{1}{|\mathcal{X}|}
\sum_{x\in\mathcal{X}}
\mathbb{I}\!\left[\mathcal{F}_{\tau}(x)\right],
\label{eq:asr}
\end{equation}

where $\mathbb{I}[\cdot]$ is the indicator function. The definition is independent
of the selected subset and failure criterion: $\mathcal{X}$ determines the
population over which ASR is computed, while $\mathcal{F}_{\tau}$ determines
which outcomes are counted as successful attacks. In our evaluation, we report
increasingly permissive criteria that consider bounding-box degradation alone,
bounding-box degradation or empty output, and additionally malformed output.

To quantify perturbation magnitude, we additionally report MS-SSIM~\cite{wang2003multiscale} for image manipulations and normalized edit distance (NED) for text manipulations. Higher MS-SSIM indicates smaller visual modifications, while lower NED indicates smaller textual modifications. MS-SSIM, and NED are reported as mean $\pm$ standard deviation over
runs categorized as \emph{Terminated with BBOX}, irrespective of whether the final IoU satisfies the ASR threshold.

ASR, MS-SSIM, and NED each capture only one facet of a test case: whether it
succeeded, and how aggressively the image or the text was altered. They are not
directly comparable across manipulation regimes, since a joint test case has both
an MS-SSIM and an NED score while a uni-modal one has only one, and neither score
alone reflects how much task degradation that aggressiveness actually bought.
Ranking test cases or comparing regimes therefore requires a single scalar that remains well-defined whether one
or multiple modalities are perturbed. To this end, we introduce the
\emph{Stealth-Weighted Task Degradation} (SWTD) metric, a modality-agnostic
aggregate that scores a test case by how much task degradation it achieves,
weighted by how unobtrusive the modifications were. Since SWTD is defined
independently of the perturbation families, the SUT, and the task oracle used
here, it can be reused as a drop-in effectiveness--stealth score by other
multi-modal robustness or test-generation methods, for example a regression suite
mixing single- and multi-modal corruptions, or a search-based generator using it
as a fitness signal, without redefining thresholds each time a new modality
combination is added. SWTD compares the task-performance
oracle $T$ (IoU in our setting) before and after manipulation with a normalized
aggressiveness measure $a^{\downarrow}_m \in [0,1]$ (lower is better, i.e.,
$a^{\downarrow}_{\operatorname{MS-SSIM}}=1-\operatorname{MS-SSIM}$) for each metric
on the manipulated modalities:

\begin{equation}
\mathrm{SWTD}
=
\frac{T_0-T_M}{T_0+\epsilon}
\left(
1-\frac{1}{|\mathcal{M}|}
\sum_{m\in\mathcal{M}} a^{\downarrow}_m
\right),
\label{eq:swtd}
\end{equation}

where $\mathcal{M}$ denotes the set of manipulated modalities and higher SWTD is
desired. Because the aggressiveness term averages over $\mathcal{M}$, SWTD reduces
to a single-modality stealth-degradation score when $|\mathcal{M}|=1$ and extends
without modification to any number of jointly manipulated modalities, which is
what allows it to rank image-only, text-only, and joint test cases on the same
scale.
Regimes are compared pairwise and paired by seed, using exact McNemar
tests~\cite{McNemar1947} for ASR and Wilcoxon signed-rank
tests~\cite{Wilcoxon1945} for MS-SSIM, NED, and SWTD, with Holm
correction~\cite{Holm1979} per metric ($\alpha=0.05$).

\subsubsection{Efficiency Metrics (RQ\textsubscript{2})}\label{sec:rq2_metrics}
Efficiency is evaluated over runs categorized as \emph{Terminated with BBOX},
irrespective of whether the final IoU satisfies the attack-success threshold.
We report the final image and text perturbation budgets, their sum $\Sigma B$,
the number of SUT evaluations, and end-to-end wall-clock runtime. One SUT evaluation corresponds to evaluating one candidate test case, and runtime includes both perturbation generation and SUT inference. Perturbation budgets lie in $[0,1]$, with larger values corresponding to stronger modifications. As for RQ\textsubscript{1}, regimes are compared with Holm-corrected Wilcoxon signed-rank tests.

\subsubsection{Validity Metrics (RQ\textsubscript{3})}
Validity is assessed through a human evaluation study on all scene types. An assessment is accepted when the perturbed image and prompt remain sufficiently understandable for the participant to perform the requested localization task. For accepted assessments, participants provide bounding boxes from which Human IoU is computed. VLM IoU is evaluated on the same accepted subset, enabling a direct comparison between human and model localization performance. Assessors may reject tasks due to 1) an unclear image, 2) an unclear prompt, or 3) another reason. 
We measure inter-rater agreement with Randolph's free-marginal multi-rater $\kappa$~\cite{randolph2005free} over two outcomes: boxes drawn or task rejected, pooling all rejection reasons into one category. We use the free-marginal variant rather than Fleiss' $\kappa$ because assessors were not told how often to pick each outcome, and because most answers are \emph{boxes drawn}. That skew pushes Fleiss' $\kappa$ down even when raters largely agree.

\subsubsection{Modality-sensitivity Metrics (RQ\textsubscript{4})}
 We characterize modality sensitivity jointly through attack effectiveness, final prediction degradation, and the perturbation budget required in each modality. A modality constitutes a stronger point of sensitivity when high attack success and large IoU reductions can be achieved with comparatively small perturbation budgets in that modality. Conversely, larger required budgets and lower attack success indicate greater robustness. Since image and text budgets parameterize different perturbation spaces, their absolute values are interpreted within their respective modalities rather than as directly equivalent quantities. SWTD, defined in \autoref{eq:swtd}, provides a complementary single-score view, reported per configuration in \autoref{tab:rq1new}.

\subsection{Optimizer Configuration}

The optimization process is performed using NSGA-II with a population size of 50 and a maximum of 100 generations. The total perturbation budget is $B=1$, shared across the active modalities. Optimization terminates early when a candidate simultaneously satisfies an IoU objective of at most $0.25$, a normalized Frobenius image distance of at most $0.1$, and a cosine embedding distance of at most $0.3$. Selection of the final returned test case follows the Pareto-optimal candidate.

The full experimental design comprises four retained VLMs, four scene types, three perturbation regimes, and 100 seed images per scene type, corresponding to 4,800 nominal optimization runs before baseline filtering and outcome-based analysis. Each run evaluates up to 5,050 ($\text{population size} \times (\text{generations} +1)$) candidate solutions in the absence of early termination.

\subsection{Data Preprocessing}
We use the ImageNet-1K~\cite{imagenet} detection validation split and the Udacity driving dataset as our image corpus. Preprocessing consists of two steps.

\subsubsection{Image selection}
Based on the ImageNet annotation structure, images are stratified into three scene types: SC-SI (single category, single instance), SC-MI (single category, multiple instances), and MC (multiple distinct object categories). We sample $N=100$ images from each stratum. In addition, we sample 100 real-world driving scenes from the Udacity dataset and use its native bounding-box annotations. All four scene types use the same structured bounding-box prompt template and have annotated ground truth bounding boxes in their respective validation split.

Baseline eligibility is determined separately for each target VLM. An image is retained for optimization only when the unperturbed model prediction achieves an IoU of at least $0.5$ against the ground-truth annotation.

\subsubsection{Semantic corruption pre-computation}
Text-based corruption operators require per-label substitution sets. We therefore collect all unique class-label strings from the selected images and query an LLM (gpt-oss:120b~\cite{agarwal2025gpt}) to generate two lookup tables: (i) homophone substitutions and (ii) synonym substitutions. These tables are computed once, cached, and reused by the corresponding corruption operators.

\subsubsection{Generation Outcomes and Model Retention}
\begin{figure}
    \centering
    \includegraphics[width=0.65\linewidth]{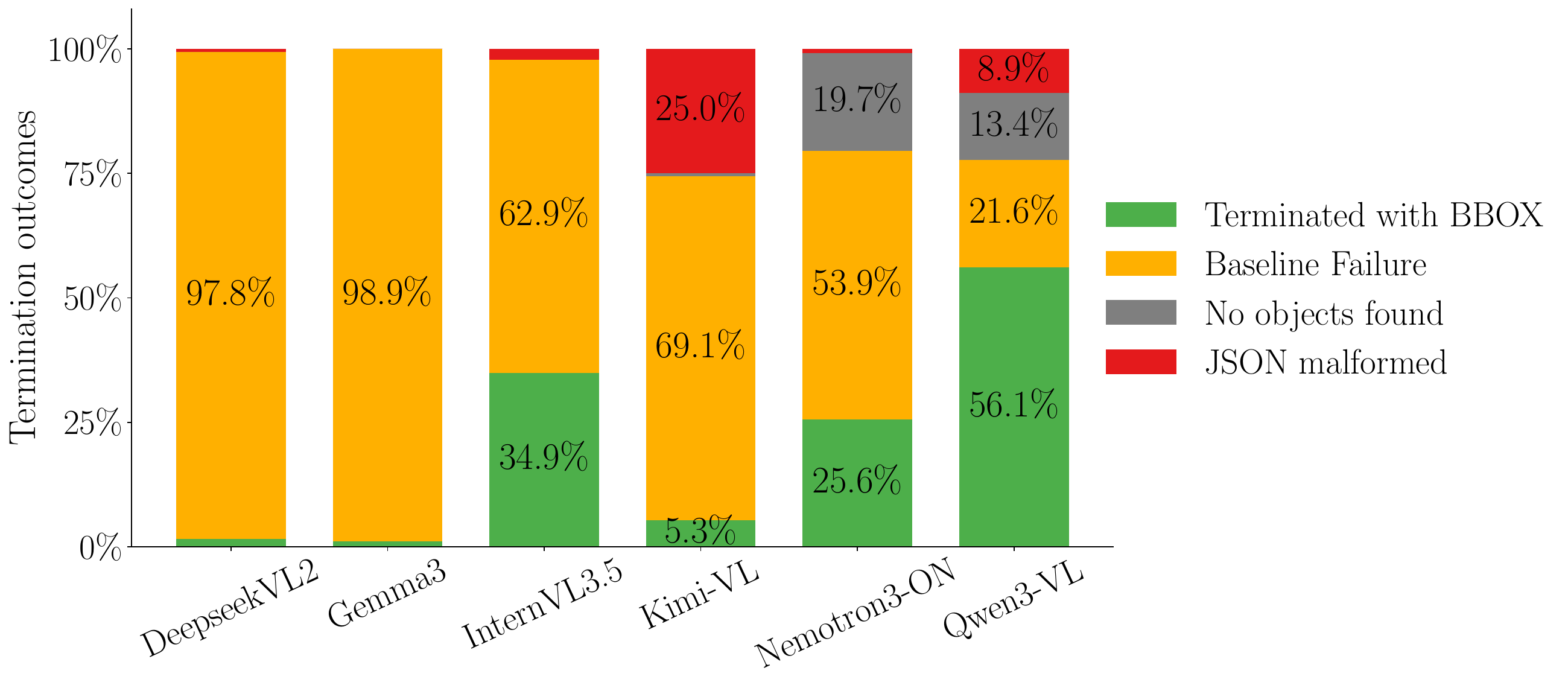}
    \caption{Distribution of test-generation run outcomes across evaluated VLMs.}
    \label{fig:termination-bars}
\end{figure}

Before the RQ-specific analyses, we categorize each test-generation run as:
(i) \emph{Baseline Failure}, when the unmodified input has IoU below $0.5$;
(ii) \emph{Terminated with BBOX}, when optimization terminates either through
early stopping or after exhausting the generation budget and the returned model
output contains valid bounding boxes;
(iii) \emph{Malformed JSON}, when the structured output cannot be parsed; or
(iv) \emph{No Objects Found}, when the SUT returns no bounding boxes.

Model retention is applied at the model level: a VLM is retained for downstream analysis if at least 5\% of all its runs result in a \emph{Terminated with BBOX} outcome. Under this criterion, InternVL, Kimi, Nemotron, and Qwen are retained. The distribution of run outcomes is shown in \autoref{fig:termination-bars}. RQ\textsubscript{1} and RQ\textsubscript{2} metric analyses are restricted to the resulting IoU-degraded test cases.
\subsection{Human Study Setup}
We constructed three human-evaluation subsets using stratified random sampling.
Each contained 48 generated test cases: 12 per VLM, with three per scene
type---one for each manipulation regime---in addition to two attention checks.
We collected at least 20 valid HITs for each subset using AWS MTurk, resulting
in 62 accepted HITs overall and $62 \times 48 = 2{,}976$ human assessments.
Workers received approximately 2€ per accepted HIT, corresponding to
approximately 124€ in direct worker compensation; responses failing the
attention checks were rejected.

For each task, participants saw the perturbed image and the exact prompt given to the VLM. They could draw zero or more bounding boxes around the requested objects or reject the task because the image was unclear, the prompt was unclear, or for another reason.

\subsection{Results}

\subsubsection{Effectiveness (RQ\textsubscript{1})}
\begin{table*}[t]
\centering
\caption{
Attack effectiveness and perturbation quality of generated test cases.
ASR$_B$ considers bounding-box (BBOX) degradation alone,
ASR$_E$ additionally counts empty output as a failure, and ASR$_M$ additionally
counts malformed output as a failure.
Bold values indicate the best result within each model and scene for the corresponding metric.
}
\label{tab:rq1new}

\setlength{\tabcolsep}{2.5pt}
\renewcommand{\arraystretch}{1.08}
\scriptsize


\begin{minipage}[t]{0.495\textwidth}
\centering
\textbf{Qwen3-VL}

\vspace{2pt}

\resizebox{\linewidth}{!}{%
\begin{tabular}{lclcccccc}
\toprule
Scene &
BL Fail &
Manip. &
ASR$_B$ &
ASR$_E$ &
ASR$_M$ &
SSIM$\uparrow$ &
NED$\downarrow$ &
SWTD$\uparrow$ \\
\midrule

\multirow{3}{*}{Driving}
& \multirow{3}{*}{0.821} & \faImage+\faFont
& 0.600
& 0.900
& \textbf{1.000}
& $\mathbf{0.957 \pm 0.038}$
& $\mathbf{0.002 \pm 0.001}$
& $\mathbf{0.884 \pm 0.191}$ \\

& & \faImage
& 0.300
& 0.300
& 0.300
& $0.954 \pm 0.035$
& ---
& $0.406 \pm 0.300$ \\

& & \faFont
& \textbf{0.700}
& \textbf{1.000}
& \textbf{1.000}
& ---
& $0.085 \pm 0.141$
& $0.848 \pm 0.258$ \\

\cmidrule(lr){1-9}

\multirow{3}{*}{MC}
& \multirow{3}{*}{0.200} & \faImage+\faFont
& \textbf{0.163}
& \textbf{0.237}
& \textbf{0.487}
& $\mathbf{0.937 \pm 0.053}$
& $\mathbf{0.211 \pm 0.169}$
& $\mathbf{0.542 \pm 0.174}$ \\

& & \faImage
& 0.025
& 0.037
& 0.225
& $0.925 \pm 0.055$
& ---
& $0.262 \pm 0.161$ \\

& & \faFont
& 0.113
& 0.175
& 0.388
& ---
& $0.277 \pm 0.126$
& $0.413 \pm 0.164$ \\

\cmidrule(lr){1-9}

\multirow{3}{*}{SC-MI}
& \multirow{3}{*}{0.060} & \faImage+\faFont
& \textbf{0.585}
& 0.745
& \textbf{0.926}
& $\mathbf{0.942 \pm 0.058}$
& $\mathbf{0.205 \pm 0.176}$
& $\mathbf{0.792 \pm 0.165}$ \\

& & \faImage
& 0.085
& 0.096
& 0.255
& $0.920 \pm 0.048$
& ---
& $0.360 \pm 0.237$ \\

& & \faFont
& 0.553
& \textbf{0.819}
& 0.883
& ---
& $0.330 \pm 0.105$
& $0.595 \pm 0.201$ \\

\cmidrule(lr){1-9}

\multirow{3}{*}{SC-SI}
& \multirow{3}{*}{0.050} & \faImage+\faFont
& \textbf{0.526}
& 0.895
& 0.926
& $\mathbf{0.958 \pm 0.050}$
& $\mathbf{0.224 \pm 0.168}$
& $\mathbf{0.813 \pm 0.159}$ \\

& & \faImage
& 0.042
& 0.053
& 0.063
& $0.906 \pm 0.054$
& ---
& $0.264 \pm 0.179$ \\

& & \faFont
& 0.432
& \textbf{0.937}
& \textbf{0.937}
& ---
& $0.352 \pm 0.100$
& $0.602 \pm 0.146$ \\

\bottomrule
\end{tabular}%
}
\end{minipage}
\hfill
\begin{minipage}[t]{0.495\textwidth}
\centering
\textbf{Kimi-VL}

\vspace{2pt}

\resizebox{\linewidth}{!}{%
\begin{tabular}{lclcccccc}
\toprule
Scene &
BL Fail &
Manip. &
ASR$_B$ &
ASR$_E$ &
ASR$_M$ &
SSIM$\uparrow$ &
NED$\downarrow$ &
SWTD$\uparrow$ \\
\midrule

\multirow{3}{*}{Driving}
& \multirow{3}{*}{0.643} & \faImage+\faFont
& \textbf{0.100}
& \textbf{0.100}
& \textbf{1.000}
& $\mathbf{0.951 \pm 0.038}$
& $\mathbf{0.033 \pm 0.103}$
& $\mathbf{0.924 \pm 0.014}$ \\

& & \faImage
& 0.050
& 0.050
& \textbf{1.000}
& $0.946 \pm 0.031$
& ---
& $0.918 \pm 0.000$ \\

& & \faFont
& 0.000
& 0.000
& \textbf{1.000}
& ---
& $0.177 \pm 0.172$
& --- \\

\cmidrule(lr){1-9}

\multirow{3}{*}{MC}
& \multirow{3}{*}{0.750} & \faImage+\faFont
& \textbf{0.160}
& \textbf{0.160}
& 0.920
& $0.903 \pm 0.072$
& $\mathbf{0.027 \pm 0.082}$
& $\mathbf{0.695 \pm 0.169}$ \\

& & \faImage
& 0.000
& 0.000
& 0.800
& $\mathbf{0.943 \pm 0.054}$
& ---
& $0.397 \pm 0.148$ \\

& & \faFont
& 0.080
& 0.080
& \textbf{0.960}
& ---
& $0.200 \pm 0.162$
& $0.644 \pm 0.205$ \\

\cmidrule(lr){1-9}

\multirow{3}{*}{SC-MI}
& \multirow{3}{*}{0.870} & \faImage+\faFont
& \textbf{0.154}
& \textbf{0.154}
& 0.692
& $0.940 \pm 0.044$
& $\mathbf{0.026 \pm 0.094}$
& $\mathbf{0.555 \pm 0.219}$ \\

& & \faImage
& \textbf{0.154}
& \textbf{0.154}
& 0.769
& $\mathbf{0.959 \pm 0.030}$
& ---
& $0.542 \pm 0.129$ \\

& & \faFont
& 0.000
& 0.000
& \textbf{0.923}
& ---
& $0.202 \pm 0.163$
& $0.224 \pm 0.000$ \\

\cmidrule(lr){1-9}

\multirow{3}{*}{SC-SI}
& \multirow{3}{*}{0.480} & \faImage+\faFont
& 0.058
& 0.077
& \textbf{0.923}
& $0.961 \pm 0.045$
& $\mathbf{0.009 \pm 0.043}$
& $\mathbf{0.715 \pm 0.212}$ \\

& & \faImage
& 0.077
& 0.115
& 0.846
& $\mathbf{0.962 \pm 0.039}$
& ---
& $0.604 \pm 0.272$ \\

& & \faFont
& \textbf{0.096}
& \textbf{0.154}
& \textbf{0.923}
& ---
& $0.118 \pm 0.126$
& $0.666 \pm 0.228$ \\

\bottomrule
\end{tabular}%
}
\end{minipage}

\vspace{6pt}


\begin{minipage}[t]{0.495\textwidth}
\centering
\textbf{InternVL3.5}

\vspace{2pt}

\resizebox{\linewidth}{!}{%
\begin{tabular}{lclcccccc}
\toprule
Scene &
BL Fail &
Manip. &
ASR$_B$ &
ASR$_E$ &
ASR$_M$ &
SSIM$\uparrow$ &
NED$\downarrow$ &
SWTD$\uparrow$ \\
\midrule

\multirow{3}{*}{Driving}
& \multirow{3}{*}{0.893} & \faImage+\faFont
& 0.667
& 0.667
& 0.833
& $0.927 \pm 0.045$
& $\mathbf{0.056 \pm 0.137}$
& $0.683 \pm 0.211$ \\

& & \faImage
& 0.833
& 0.833
& \textbf{1.000}
& $\mathbf{0.945 \pm 0.033}$
& ---
& $0.728 \pm 0.184$ \\

& & \faFont
& \textbf{1.000}
& \textbf{1.000}
& \textbf{1.000}
& ---
& $0.225 \pm 0.163$
& $\mathbf{0.750 \pm 0.144}$ \\

\cmidrule(lr){1-9}

\multirow{3}{*}{MC}
& \multirow{3}{*}{0.480} & \faImage+\faFont
& \textbf{0.731}
& \textbf{0.731}
& \textbf{0.846}
& $0.894 \pm 0.059$
& $\mathbf{0.052 \pm 0.109}$
& $\mathbf{0.708 \pm 0.147}$ \\

& & \faImage
& 0.654
& 0.654
& 0.769
& $\mathbf{0.915 \pm 0.052}$
& ---
& $0.666 \pm 0.157$ \\

& & \faFont
& 0.712
& 0.712
& 0.808
& ---
& $0.251 \pm 0.145$
& $0.552 \pm 0.144$ \\

\cmidrule(lr){1-9}

\multirow{3}{*}{SC-MI}
& \multirow{3}{*}{0.580} & \faImage+\faFont
& 0.905
& 0.905
& \textbf{0.952}
& $0.904 \pm 0.059$
& $\mathbf{0.084 \pm 0.141}$
& $\mathbf{0.787 \pm 0.132}$ \\

& & \faImage
& 0.833
& 0.833
& 0.857
& $\mathbf{0.906 \pm 0.050}$
& ---
& $0.762 \pm 0.143$ \\

& & \faFont
& \textbf{0.929}
& \textbf{0.929}
& 0.929
& ---
& $0.324 \pm 0.119$
& $0.584 \pm 0.118$ \\

\cmidrule(lr){1-9}

\multirow{3}{*}{SC-SI}
& \multirow{3}{*}{0.680} & \faImage+\faFont
& \textbf{0.969}
& \textbf{0.969}
& \textbf{0.969}
& $\mathbf{0.907 \pm 0.065}$
& $\mathbf{0.138 \pm 0.163}$
& $\mathbf{0.753 \pm 0.123}$ \\

& & \faImage
& 0.781
& 0.781
& 0.781
& $0.901 \pm 0.060$
& ---
& $0.724 \pm 0.166$ \\

& & \faFont
& 0.875
& 0.875
& 0.906
& ---
& $0.249 \pm 0.137$
& $0.642 \pm 0.185$ \\

\bottomrule
\end{tabular}%
}
\end{minipage}
\hfill
\begin{minipage}[t]{0.495\textwidth}
\centering
\textbf{Nemotron3-ON}

\vspace{2pt}

\resizebox{\linewidth}{!}{%
\begin{tabular}{lclcccccc}
\toprule
Scene &
BL Fail &
Manip. &
ASR$_B$ &
ASR$_E$ &
ASR$_M$ &
SSIM$\uparrow$ &
NED$\downarrow$ &
SWTD$\uparrow$ \\
\midrule

\multirow{3}{*}{Driving}
& \multirow{3}{*}{0.732} & \faImage+\faFont
& \textbf{0.600}
& \textbf{1.000}
& \textbf{1.000}
& $\mathbf{0.943 \pm 0.038}$
& $\mathbf{0.155 \pm 0.172}$
& $\mathbf{0.795 \pm 0.163}$ \\

& & \faImage
& \textbf{0.600}
& 0.667
& 0.667
& $0.927 \pm 0.040$
& ---
& $0.575 \pm 0.271$ \\

& & \faFont
& 0.200
& 0.933
& 0.933
& ---
& $0.221 \pm 0.204$
& $0.675 \pm 0.272$ \\

\cmidrule(lr){1-9}

\multirow{3}{*}{MC}
& \multirow{3}{*}{0.590} & \faImage+\faFont
& \textbf{0.317}
& 0.415
& 0.439
& $\mathbf{0.908 \pm 0.052}$
& $\mathbf{0.011 \pm 0.012}$
& $0.636 \pm 0.195$ \\

& & \faImage
& 0.293
& 0.293
& 0.317
& $0.880 \pm 0.065$
& ---
& $0.466 \pm 0.178$ \\

& & \faFont
& 0.220
& \textbf{0.439}
& \textbf{0.488}
& ---
& $0.083 \pm 0.117$
& $\mathbf{0.654 \pm 0.223}$ \\

\cmidrule(lr){1-9}

\multirow{3}{*}{SC-MI}
& \multirow{3}{*}{0.880} & \faImage+\faFont
& \textbf{0.417}
& \textbf{0.917}
& \textbf{0.917}
& $\mathbf{0.912 \pm 0.064}$
& $\mathbf{0.013 \pm 0.028}$
& $\mathbf{0.831 \pm 0.263}$ \\

& & \faImage
& 0.250
& 0.500
& 0.500
& $0.893 \pm 0.050$
& ---
& $0.570 \pm 0.283$ \\

& & \faFont
& 0.333
& \textbf{0.917}
& \textbf{0.917}
& ---
& $0.260 \pm 0.130$
& $0.652 \pm 0.231$ \\

\cmidrule(lr){1-9}

\multirow{3}{*}{SC-SI}
& \multirow{3}{*}{0.040} & \faImage+\faFont
& 0.104
& 0.844
& 0.844
& $\mathbf{0.918 \pm 0.058}$
& $\mathbf{0.082 \pm 0.138}$
& $\mathbf{0.845 \pm 0.185}$ \\

& & \faImage
& \textbf{0.135}
& 0.198
& 0.208
& $0.871 \pm 0.052$
& ---
& $0.469 \pm 0.206$ \\

& & \faFont
& 0.062
& \textbf{0.958}
& \textbf{1.000}
& ---
& $0.318 \pm 0.089$
& $0.678 \pm 0.085$ \\

\bottomrule
\end{tabular}%
}
\end{minipage}

\end{table*}
\begin{figure}
    \centering
    \includegraphics[width=0.8\linewidth]{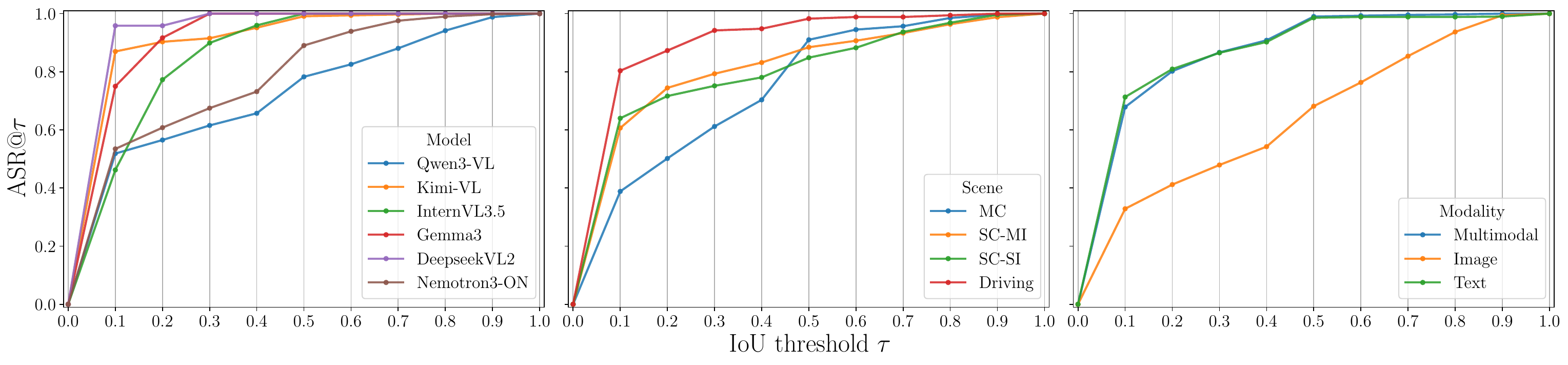}
    \caption{ASR as a function over threshold $\tau$, grouped by either Model, Scene or Manipulation Strategy.}
    \label{fig:rq1_asr}
\end{figure}
We evaluate effectiveness using three increasingly permissive attack-success
criteria: BBOX degradation below $\tau=0.25$, BBOX or empty output, and BBOX,
empty, or malformed output. \autoref{tab:rq1new} reports these success rates
together with MS-SSIM, NED, and SWTD, while \autoref{fig:rq1_asr} shows the
sensitivity of ASR to the selected IoU threshold.

Effectiveness depends heavily on the model and the regime.
Qwen shows a clear susceptibility to textual perturbations, although multi-modal
manipulation is strongest overall on MC and SC-MI, and ties with text-only on
SC-SI. InternVL exhibits high direct localization degradation, with text-only
performing best on Driving, multi-modal manipulation dominating MC and SC-SI,
and the two tied on SC-MI. Nemotron is mixed, with multi-modal
manipulation winning on Driving and SC-MI, and text-only manipulation winning
on MC and SC-SI. Kimi exhibits comparatively low strict BBOX
ASRs, with multi-modal manipulation strongest overall on Driving, MC, and
SC-MI, and text-only strongest on SC-SI.

The permissive criteria catch failures that the IoU alone misses, most visibly
for Kimi:
for example, multi-modal ASR on Driving increases from $0.1$ under BBOX to
$1$ when malformed outputs are included, while MC increases from $0.16$
to $0.92$. Since the intermediate BBOX+Empty values remain unchanged in these
cases, these increases are primarily caused by malformed structured outputs.
Thus, high permissive ASR can reflect output-format failures in addition to
localization degradation.

Only the ASR variants and SWTD are defined for all three regimes (MS-SSIM and
NED each cover one modality). Taken at face value, multi-modal manipulation has
the best or tied-best score in 11 of the 16 model--scene configurations and
text-only in the other five. The paired tests (\autoref{sec:rq1_metrics}) show that many of these margins
are small, but they point to a consistent picture.

For attack success, the significant McNemar differences all separate
manipulating the prompt from manipulating only the image. Joint and text-only
manipulation significantly outperform image-only for Qwen on the ImageNet-derived
scenes and, once empty outputs count as failures, for Nemotron on SC-SI.
Image-only manipulation is never significantly more effective than either
alternative. Joint and text-only manipulation, in turn, are statistically
indistinguishable in every configuration but one (Nemotron SC-SI under ASR$_M$,
where text-only is stronger). For Kimi and InternVL, no regime separates from
the others.

The clearer advantage of joint manipulation lies
in how failures are reached rather than how often. Joint test cases require
significantly smaller textual edits than text-only ones (lower NED, Wilcoxon) in 12 of the
16 configurations, covering every model on every ImageNet-derived scene except
Qwen MC. This reduction comes without a loss in visual fidelity: MS-SSIM of
joint test cases is on par with image-only in most configurations, significantly
higher for Qwen (SC-MI, SC-SI) and Nemotron SC-SI, and significantly lower only
for InternVL MC. SWTD, which weights degradation by how unobtrusive the
modifications are, consequently favors joint manipulation significantly over
image-only in five configurations and over text-only in six, mostly for Qwen,
InternVL, and Nemotron on the ImageNet-derived scenes, while joint manipulation
is never significantly worse than either uni-modal regime. For Kimi, SWTD
comparisons are largely uninformative, as most of its failures are malformed
outputs for which SWTD is undefined.

\begin{tcolorbox}[boxrule=0pt,sharp corners,boxsep=2pt,left=2pt,right=2pt,top=2.5pt,bottom=2pt]
\textbf{RQ\textsubscript{1} (Effectiveness):}
\textit{
\tool induces both localization and structured-output failures across all evaluated VLMs. Manipulating the prompt, alone or jointly with the image, is significantly more effective than image-only manipulation for Qwen and Nemotron, and image-only is never significantly best. Joint manipulation matches text-only in attack success while requiring significantly smaller textual edits in 12 of 16 configurations, and it is never significantly worse in SWTD. Kimi additionally shows a pronounced tendency toward malformed outputs.
}
\end{tcolorbox}

\subsubsection{Efficiency (RQ\textsubscript{2})}
\begin{table*}[t]
\centering
\caption{
Efficiency results. Values are mean $\pm$ standard deviation; bold marks the lowest
value per model--scene block for each efficiency metric.
Green indicates the RQ\textsubscript{1}-winning manipulation(s) from \autoref{tab:rq1new}.
}
\label{tab:efficiency}

\setlength{\tabcolsep}{2.5pt}
\renewcommand{\arraystretch}{1.08}
\scriptsize


\begin{minipage}[t]{0.495\textwidth}
\centering
\textbf{Qwen3-VL}

\vspace{2pt}

\resizebox{\linewidth}{!}{%
\begin{tabular}{llccccc}
\toprule
Scene &
Manip. &
$B_{\mathrm{img}}\downarrow$ &
$B_{\mathrm{txt}}\downarrow$ &
$\Sigma B\downarrow$ &
\#Eval$\downarrow$ &
Time (s)$\downarrow$ \\
\midrule


\multirow{3}{*}{Driving}
& \faImage+\faFont
& $0.174 \pm 0.120$
& $\mathbf{0.141 \pm 0.050}$
& $0.315 \pm 0.152$
& $330 \pm 414$
& $16.55 \pm 19.82$
\\

& \faImage
& $\mathbf{0.120 \pm 0.089}$
& ---
& $\mathbf{0.120 \pm 0.089}$
& $3520 \pm 2383$
& $109.11 \pm 69.79$
\\

& \faFont
& \cellcolor{green!15}---
& \cellcolor{green!15}$0.465 \pm 0.190$
& \cellcolor{green!15}$0.465 \pm 0.190$
& \cellcolor{green!15}$\mathbf{175 \pm 132}$
& \cellcolor{green!15}$\mathbf{5.06 \pm 3.81}$
\\

\cmidrule(lr){1-7}

\multirow{3}{*}{MC}
& \faImage+\faFont
& \cellcolor{green!15}$\mathbf{0.196 \pm 0.150}$
& \cellcolor{green!15}$\mathbf{0.405 \pm 0.161}$
& \cellcolor{green!15}$0.601 \pm 0.218$
& \cellcolor{green!15}$\mathbf{458 \pm 1080}$
& \cellcolor{green!15}$33.93 \pm 62.95$
\\

& \faImage
& $0.219 \pm 0.136$
& ---
& $\mathbf{0.219 \pm 0.136}$
& $3003 \pm 2349$
& $193.70 \pm 162.88$
\\

& \faFont
& ---
& $0.439 \pm 0.149$
& $0.439 \pm 0.149$
& $558 \pm 1194$
& $\mathbf{30.00 \pm 49.84}$
\\

\cmidrule(lr){1-7}


\multirow{3}{*}{SC-MI}
& \faImage+\faFont
& \cellcolor{green!15}$\mathbf{0.186 \pm 0.149}$
& \cellcolor{green!15}$\mathbf{0.440 \pm 0.162}$
& \cellcolor{green!15}$0.626 \pm 0.217$
& \cellcolor{green!15}$\mathbf{812 \pm 1363}$
& \cellcolor{green!15}$61.05 \pm 92.72$
\\

& \faImage
& $0.206 \pm 0.120$
& ---
& $\mathbf{0.206 \pm 0.120}$
& $2996 \pm 2340$
& $158.31 \pm 137.09$
\\

& \faFont
& ---
& $0.513 \pm 0.195$
& $0.513 \pm 0.195$
& $1374 \pm 1916$
& $\mathbf{54.81 \pm 79.86}$
\\

\cmidrule(lr){1-7}


\multirow{3}{*}{SC-SI}
& \faImage+\faFont
& \cellcolor{green!15}$\mathbf{0.136 \pm 0.143}$
& \cellcolor{green!15}$\mathbf{0.533 \pm 0.179}$
& \cellcolor{green!15}$0.668 \pm 0.201$
& \cellcolor{green!15}$\mathbf{1095 \pm 1596}$
& \cellcolor{green!15}$63.23 \pm 92.97$
\\

& \faImage
& $0.247 \pm 0.114$
& ---
& $\mathbf{0.247 \pm 0.114}$
& $4489 \pm 1420$
& $168.04 \pm 55.89$
\\

& \faFont
& \cellcolor{green!15}---
& \cellcolor{green!15}$0.613 \pm 0.228$
& \cellcolor{green!15}$0.613 \pm 0.228$
& \cellcolor{green!15}$1327 \pm 1833$
& \cellcolor{green!15}$\mathbf{34.37 \pm 48.12}$
\\

\bottomrule
\end{tabular}%
}
\end{minipage}
\hfill
\begin{minipage}[t]{0.495\textwidth}
\centering
\textbf{Kimi-VL}

\vspace{2pt}

\resizebox{\linewidth}{!}{%
\begin{tabular}{llccccc}
\toprule
Scene &
Manip. &
$B_{\mathrm{img}}\downarrow$ &
$B_{\mathrm{txt}}\downarrow$ &
$\Sigma B\downarrow$ &
\#Eval$\downarrow$ &
Time (s)$\downarrow$ \\
\midrule


\multirow{3}{*}{Driving}
& \faImage+\faFont
& \cellcolor{green!15}$\mathbf{0.110 \pm 0.064}$
& \cellcolor{green!15}$\mathbf{0.082 \pm 0.053}$
& \cellcolor{green!15}$0.192 \pm 0.094$
& \cellcolor{green!15}$55 \pm 22$
& \cellcolor{green!15}$36.42 \pm 29.46$
\\

& \faImage
& $0.156 \pm 0.076$
& ---
& $\mathbf{0.156 \pm 0.076}$
& $58 \pm 24$
& $31.09 \pm 29.73$
\\

& \faFont
& ---
& $0.646 \pm 0.242$
& $0.646 \pm 0.242$
& $\mathbf{53 \pm 11}$
& $\mathbf{16.83 \pm 25.26}$
\\

\cmidrule(lr){1-7}


\multirow{3}{*}{MC}
& \faImage+\faFont
& \cellcolor{green!15}$\mathbf{0.195 \pm 0.147}$
& \cellcolor{green!15}$\mathbf{0.180 \pm 0.131}$
& \cellcolor{green!15}$0.375 \pm 0.220$
& \cellcolor{green!15}$54 \pm 14$
& \cellcolor{green!15}$\mathbf{7.45 \pm 3.09}$
\\

& \faImage
& $0.268 \pm 0.230$
& ---
& $\mathbf{0.268 \pm 0.230}$
& $328 \pm 1023$
& $50.55 \pm 168.67$
\\

& \faFont
& ---
& $0.562 \pm 0.287$
& $0.562 \pm 0.287$
& $\mathbf{50 \pm 0}$
& $8.77 \pm 16.09$
\\

\cmidrule(lr){1-7}


\multirow{3}{*}{SC-MI}
& \faImage+\faFont
& \cellcolor{green!15}$\mathbf{0.134 \pm 0.098}$
& \cellcolor{green!15}$\mathbf{0.181 \pm 0.127}$
& \cellcolor{green!15}$0.314 \pm 0.191$
& \cellcolor{green!15}$81 \pm 85$
& \cellcolor{green!15}$8.58 \pm 7.55$
\\

& \faImage
& $0.235 \pm 0.185$
& ---
& $\mathbf{0.235 \pm 0.185}$
& $\mathbf{77 \pm 73}$
& $\mathbf{6.19 \pm 5.01}$
\\

& \faFont
& ---
& $0.634 \pm 0.285$
& $0.634 \pm 0.285$
& $204 \pm 540$
& $25.53 \pm 66.86$
\\

\cmidrule(lr){1-7}


\multirow{3}{*}{SC-SI}
& \faImage+\faFont
& $\mathbf{0.070 \pm 0.066}$
& $\mathbf{0.090 \pm 0.080}$
& $0.160 \pm 0.119$
& $325 \pm 843$
& $18.05 \pm 45.13$
\\

& \faImage
& $0.081 \pm 0.056$
& ---
& $\mathbf{0.081 \pm 0.056}$
& $637 \pm 1447$
& $29.43 \pm 66.18$
\\

& \faFont
& \cellcolor{green!15}---
& \cellcolor{green!15}$0.529 \pm 0.240$
& \cellcolor{green!15}$0.529 \pm 0.240$
& \cellcolor{green!15}$\mathbf{185 \pm 323}$
& \cellcolor{green!15}$\mathbf{10.78 \pm 18.06}$
\\

\bottomrule
\end{tabular}%
}
\end{minipage}

\vspace{6pt}


\begin{minipage}[t]{0.495\textwidth}
\centering
\textbf{InternVL3.5}

\vspace{2pt}

\resizebox{\linewidth}{!}{%
\begin{tabular}{llccccc}
\toprule
Scene &
Manip. &
$B_{\mathrm{img}}\downarrow$ &
$B_{\mathrm{txt}}\downarrow$ &
$\Sigma B\downarrow$ &
\#Eval$\downarrow$ &
Time (s)$\downarrow$ \\
\midrule


\multirow{3}{*}{Driving}
& \faImage+\faFont
& $0.147 \pm 0.160$
& $\mathbf{0.085 \pm 0.089}$
& $0.232 \pm 0.249$
& $150 \pm 221$
& $45.35 \pm 48.28$
\\

& \faImage
& $\mathbf{0.128 \pm 0.049}$
& ---
& $\mathbf{0.128 \pm 0.049}$
& $333 \pm 580$
& $72.31 \pm 89.67$
\\

& \faFont
& \cellcolor{green!15}---
& \cellcolor{green!15}$0.619 \pm 0.129$
& \cellcolor{green!15}$0.619 \pm 0.129$
& \cellcolor{green!15}$\mathbf{58 \pm 20}$
& \cellcolor{green!15}$\mathbf{11.44 \pm 19.20}$
\\

\cmidrule(lr){1-7}


\multirow{3}{*}{MC}
& \faImage+\faFont
& \cellcolor{green!15}$0.301 \pm 0.176$
& \cellcolor{green!15}$\mathbf{0.195 \pm 0.132}$
& \cellcolor{green!15}$0.496 \pm 0.281$
& \cellcolor{green!15}$63 \pm 45$
& \cellcolor{green!15}$29.01 \pm 23.54$
\\

& \faImage
& $\mathbf{0.298 \pm 0.170}$
& ---
& $\mathbf{0.298 \pm 0.170}$
& $217 \pm 720$
& $75.88 \pm 194.79$
\\

& \faFont
& ---
& $0.587 \pm 0.224$
& $0.587 \pm 0.224$
& $\mathbf{55 \pm 18}$
& $\mathbf{7.11 \pm 9.65}$
\\

\cmidrule(lr){1-7}

\multirow{3}{*}{SC-MI}
& \faImage+\faFont
& \cellcolor{green!15}$\mathbf{0.253 \pm 0.145}$
& \cellcolor{green!15}$\mathbf{0.206 \pm 0.141}$
& \cellcolor{green!15}$0.459 \pm 0.240$
& \cellcolor{green!15}$75 \pm 110$
& \cellcolor{green!15}$29.99 \pm 53.48$
\\

& \faImage
& $0.272 \pm 0.118$
& ---
& $\mathbf{0.272 \pm 0.118}$
& $249 \pm 890$
& $105.53 \pm 395.83$
\\

& \faFont
& \cellcolor{green!15}---
& \cellcolor{green!15}$0.671 \pm 0.176$
& \cellcolor{green!15}$0.671 \pm 0.176$
& \cellcolor{green!15}$\mathbf{58 \pm 38}$
& \cellcolor{green!15}$\mathbf{6.29 \pm 8.69}$
\\

\cmidrule(lr){1-7}


\multirow{3}{*}{SC-SI}
& \faImage+\faFont
& \cellcolor{green!15}$\mathbf{0.233 \pm 0.133}$
& \cellcolor{green!15}$\mathbf{0.226 \pm 0.165}$
& \cellcolor{green!15}$0.459 \pm 0.260$
& \cellcolor{green!15}$84 \pm 75$
& \cellcolor{green!15}$26.64 \pm 28.49$
\\

& \faImage
& $0.269 \pm 0.127$
& ---
& $\mathbf{0.269 \pm 0.127}$
& $258 \pm 886$
& $92.26 \pm 345.47$
\\

& \faFont
& ---
& $0.654 \pm 0.183$
& $0.654 \pm 0.183$
& $\mathbf{67 \pm 89}$
& $\mathbf{6.07 \pm 11.84}$
\\

\bottomrule
\end{tabular}%
}
\end{minipage}
\hfill
\begin{minipage}[t]{0.495\textwidth}
\centering
\textbf{Nemotron3-ON}

\vspace{2pt}

\resizebox{\linewidth}{!}{%
\begin{tabular}{llccccc}
\toprule
Scene &
Manip. &
$B_{\mathrm{img}}\downarrow$ &
$B_{\mathrm{txt}}\downarrow$ &
$\Sigma B\downarrow$ &
\#Eval$\downarrow$ &
Time (s)$\downarrow$ \\
\midrule


\multirow{3}{*}{Driving}
& \faImage+\faFont
& \cellcolor{green!15}$\mathbf{0.315 \pm 0.219}$
& \cellcolor{green!15}$\mathbf{0.342 \pm 0.111}$
& \cellcolor{green!15}$0.657 \pm 0.202$
& \cellcolor{green!15}$\mathbf{320 \pm 787}$
& \cellcolor{green!15}$26.97 \pm 68.22$
\\

& \faImage
& $0.449 \pm 0.238$
& ---
& $\mathbf{0.449 \pm 0.238}$
& $1880 \pm 2355$
& $153.67 \pm 199.00$
\\

& \faFont
& ---
& $0.708 \pm 0.224$
& $0.708 \pm 0.224$
& $387 \pm 736$
& $\mathbf{22.93 \pm 37.31}$
\\

\cmidrule(lr){1-7}


\multirow{3}{*}{MC}
& \faImage+\faFont
& $\mathbf{0.188 \pm 0.092}$
& $\mathbf{0.196 \pm 0.083}$
& $\mathbf{0.383 \pm 0.155}$
& $62 \pm 35$
& $9.70 \pm 10.91$
\\

& \faImage
& $0.780 \pm 0.282$
& ---
& $0.780 \pm 0.282$
& $1354 \pm 2119$
& $117.47 \pm 180.09$
\\

& \faFont
& \cellcolor{green!15}---
& \cellcolor{green!15}$0.563 \pm 0.248$
& \cellcolor{green!15}$0.563 \pm 0.248$
& \cellcolor{green!15}$\mathbf{50 \pm 0}$
& \cellcolor{green!15}$\mathbf{6.62 \pm 9.55}$
\\

\cmidrule(lr){1-7}


\multirow{3}{*}{SC-MI}
& \faImage+\faFont
& \cellcolor{green!15}$\mathbf{0.242 \pm 0.196}$
& \cellcolor{green!15}$\mathbf{0.155 \pm 0.069}$
& \cellcolor{green!15}$\mathbf{0.397 \pm 0.241}$
& \cellcolor{green!15}$688 \pm 1392$
& \cellcolor{green!15}$72.08 \pm 161.49$
\\

& \faImage
& $0.815 \pm 0.242$
& ---
& $0.815 \pm 0.242$
& $2408 \pm 2493$
& $216.36 \pm 259.48$
\\

& \faFont
& ---
& $0.516 \pm 0.146$
& $0.516 \pm 0.146$
& $\mathbf{567 \pm 1404}$
& $\mathbf{39.67 \pm 103.22}$
\\

\cmidrule(lr){1-7}


\multirow{3}{*}{SC-SI}
& \faImage+\faFont
& $\mathbf{0.152 \pm 0.095}$
& $\mathbf{0.190 \pm 0.113}$
& $\mathbf{0.342 \pm 0.165}$
& $1051 \pm 1575$
& $69.79 \pm 105.28$
\\

& \faImage
& $0.863 \pm 0.179$
& ---
& $0.863 \pm 0.179$
& $2679 \pm 2337$
& $168.33 \pm 147.19$
\\

& \faFont
& \cellcolor{green!15}---
& \cellcolor{green!15}$0.604 \pm 0.148$
& \cellcolor{green!15}$0.604 \pm 0.148$
& \cellcolor{green!15}$\mathbf{538 \pm 863}$
& \cellcolor{green!15}$\mathbf{29.08 \pm 46.10}$
\\

\bottomrule
\end{tabular}%
}
\end{minipage}

\end{table*}

\autoref{tab:efficiency} reports perturbation budgets, SUT evaluations, and end-to-end runtime for IoU-degraded test cases. Multi-modal perturbations require smaller modality-specific budgets than their uni-modal counterparts, and this effect is most pronounced on the text side: according to the Wilcoxon signed-rank tests (\autoref{sec:rq2_metrics}), joint test cases use a significantly smaller text budget than text-only ones in 13 of the 16 configurations. The image-budget reduction relative to image-only is significant in only four configurations, three of them for Nemotron, whose image-only attacks require unusually large budgets. Because joint perturbations distribute modifications across both modalities, their aggregate budget $\Sigma B$ is typically higher than image-only and lower than or on par with text-only perturbation. Since the two modality budgets parameterize different perturbation spaces, $\Sigma B$ is treated only as a compact optimizer-space summary.

Computational effort is dominated by the image modality. For Qwen and Nemotron, image-only search requires significantly more SUT evaluations than both text-only and joint search on most ImageNet-derived scenes, and image-only search is significantly slower than text-only search in 10 of the 16 configurations. The paired tests detect no statistically significant difference between joint and text-only manipulation in all but one configuration. Text-only search is nonetheless significantly faster in wall-clock time for InternVL on all ImageNet-derived scenes and for Nemotron and Qwen on selected scenes. This suggests that differences in per-candidate processing cost, rather than evaluation counts alone, contribute to the runtime gap. For Kimi, differences between regimes are small and mostly not significant.

Driving scenes do not consistently require more search effort than the ImageNet-derived scenes. Instead, computational efficiency depends primarily on the target model and manipulation regime.

\begin{tcolorbox}[boxrule=0pt,sharp corners,boxsep=2pt,left=2pt,right=2pt,top=2.5pt,bottom=2pt]
\begin{center}
\begin{minipage}[t]{0.99\linewidth}
\textbf{RQ\textsubscript{2} (Efficiency):}
\textit{Efficiency is model- and modality-dependent. Image-only search is the most expensive regime, while text-only and joint search need a comparable number of SUT evaluations, with text-only search usually fastest in wall-clock time; for Kimi, no regime is consistently cheaper. Joint search significantly reduces the text budget relative to text-only in 13 of 16 configurations, but its aggregate budget is not consistently lower.}
\end{minipage}
\end{center}
\end{tcolorbox}

\subsubsection{Validity (RQ\textsubscript{3})}

\begin{figure}
    \centering
    \includegraphics[width=0.7\linewidth]{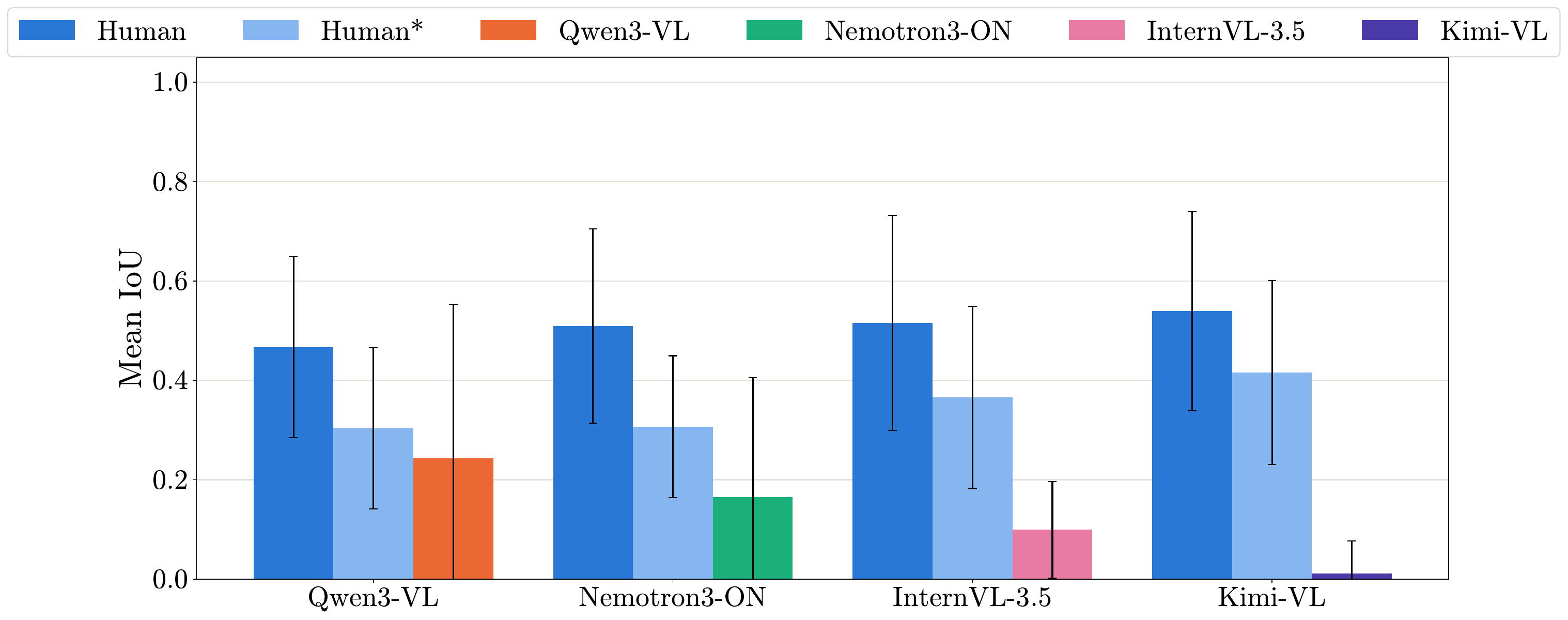}
    \caption{
    Human and VLM localization performance by target model, pooled across
    manipulation and scene types.
    }
    \label{fig:rq3_val_fig}
\end{figure}

We assess the validity of the generated test cases from two complementary
perspectives: whether human assessors consider a case interpretable, and whether
the requested object remains localizable to humans when the corresponding VLM
fails. Human acceptance
rates range from 60.2\% to 75.9\%, with the highest acceptance for Kimi-VL
(75.9\%), followed by InternVL3.5 (70.6\%), Qwen3-VL (63.3\%), and
Nemotron3-ON (60.2\%). For every evaluated model, a majority of human assessments
judged the perturbed inputs sufficiently interpretable to perform the localization task.

The rejection categories provide additional insight into why cases are considered
invalid. For InternVL3.5 and Kimi-VL, unclear images account for 14.0\% and
10.7\% of assessments, while unclear prompts account for 11.6\% and 8.7\%,
respectively. For Nemotron3-ON, image- and prompt-related rejection rates are
similar at 16.4\% and 17.0\%. Qwen3-VL exhibits a different pattern: only
11.5\% of assessments reject a case because of an unclear image, whereas
21.6\% reject it because of an unclear prompt. Rejections for other reasons
remain comparatively uncommon across all models, ranging from 3.6\% to 6.4\%.
Overall, invalid cases are therefore primarily associated with ambiguity in the
generated image or prompt rather than other failure modes.

To determine whether cases that fool the VLMs nevertheless remain understandable
to humans, we compare localization performance in
\autoref{fig:rq3_val_fig}. \emph{Human} denotes the mean IoU among accepting
human assessors. \emph{Human*} is more conservative: non-accepted assessments
are assigned an IoU of zero, thereby combining interpretability and localization
performance into a single measure. The VLM score reports the localization IoU
of the corresponding target model. The comparison is pooled across manipulation
and scene types, with 36 generated cases per model.

On accepted cases, humans consistently localize the requested object substantially
more accurately than the corresponding VLM. Mean human IoU is $0.467$ for
Qwen3-VL, $0.51$ for Nemotron3-ON, $0.516$ for InternVL3.5, and $0.540$
for Kimi-VL. In comparison, the corresponding model IoUs are only $0.243$,
$0.165$, $0.099$, and $0.012$, respectively. The largest separation occurs
for Kimi-VL, where human localization remains high despite almost complete
localization failure by the model.

The stricter Human* measure gives the same picture.
Human* IoU is $0.303$ for Qwen3-VL, $0.307$ for Nemotron3-ON, $0.366$
for InternVL3.5, and $0.416$ for Kimi-VL. These values remain higher than
the corresponding VLM IoUs in every case. The gap is therefore not an artifact of dropping rejected cases: even when
every rejection counts as a miss, humans still localize the objects better than
the models the tests were built against.

Agreement between assessors is only slight to fair. Randolph's $\kappa$ is
$0.2$ overall, highest for Driving ($0.244$) and SC-SI ($0.229$), and lower
for SC-MI ($0.172$) and MC ($0.155$). The ordering makes sense: a single object
in a clear scene is easy to judge, while a cluttered multi-category image leaves
more room to argue about whether a blurred or partly occluded object is still
recognizable. 
Agreement on interpretability is limited
(Randolph's $\kappa=0.200$). We therefore report both acceptance
rates and Human*, which assigns zero IoU to rejected assessments.

\begin{tcolorbox}[boxrule=0pt,sharp corners,boxsep=2pt,left=2pt,right=2pt,top=2.5pt,bottom=2pt]
\begin{center}
\begin{minipage}[t]{0.99\linewidth}
\textbf{RQ\textsubscript{3} (Validity):}
\textit{
A majority of generated test cases remain interpretable, with acceptance rates between 60.2\% and 75.9\% across models. Humans localize the requested object substantially more accurately than the corresponding VLM (accepted-case IoU $0.467$--$0.540$ vs.\ $0.012$--$0.243$), and this gap persists even under the conservative Human* measure. Agreement between assessors is only slight to fair (Randolph's $\kappa=0.2$), so validity is not always clear-cut.
}
\end{minipage}
\end{center}
\end{tcolorbox}
\subsubsection{Modality Sensitivity (RQ\textsubscript{4})}\label{sec:mod_sens}

\begin{figure}[tb]
    \centering
    \includegraphics[width=0.92\linewidth]{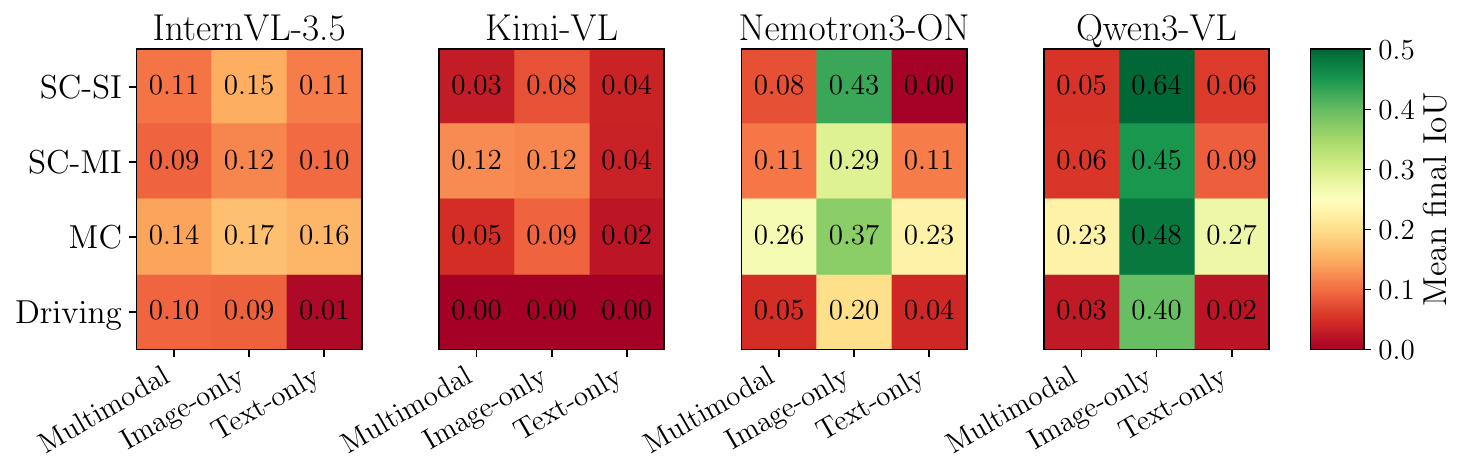}
    \caption{Mean final IoU by model, scene type, and manipulation modality.}
    \label{fig:smheat}
\end{figure}

\autoref{fig:smheat} shows the mean final IoU by model, scene type, and manipulation modality. The resulting profiles differ substantially across VLMs. Qwen exhibits the clearest modality imbalance: image-only manipulation consistently retains much higher final IoU than text-only or multi-modal manipulation, indicating greater robustness to visual than textual perturbations. Nemotron shows a similar, though less uniform, pattern, particularly for SC-SI and Driving scenes. In contrast, Kimi reaches low final IoU under all manipulation regimes, while InternVL exhibits comparatively smaller differences between modalities.

\begin{figure}[tb]
    \centering
    \includegraphics[width=0.92\linewidth]{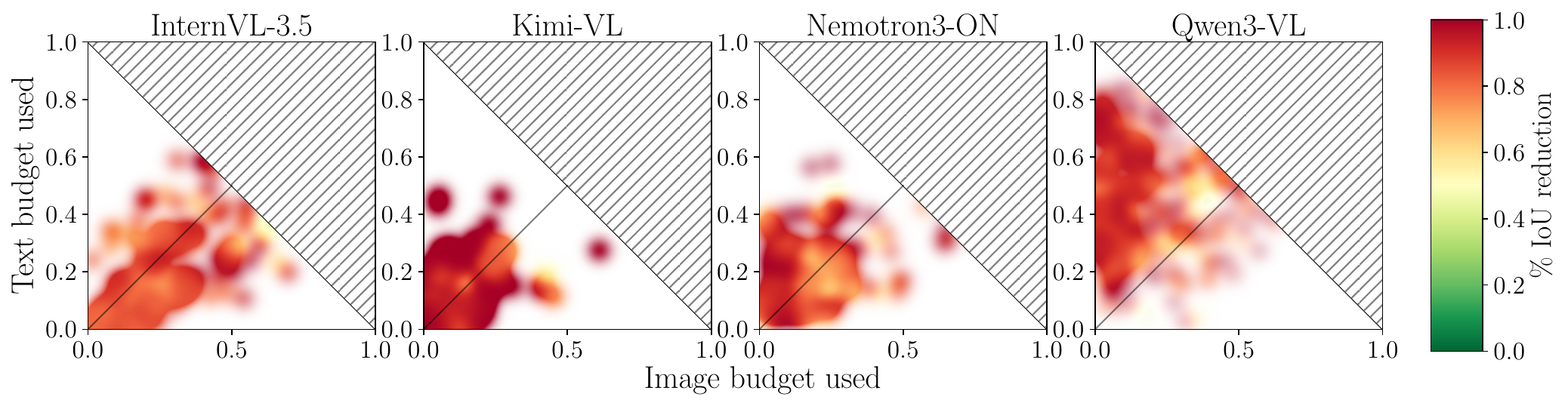}
    \caption{
IoU percentage reduction as a function of image and text perturbation budgets
for multi-modal test cases. The hatched region denotes infeasible allocations
for which $B_{\mathrm{img}}+B_{\mathrm{txt}}>1$.
}
    \label{fig:ioured}
\end{figure}

The budget-allocation analysis in \autoref{fig:ioured} further distinguishes these profiles. Qwen's strongest IoU reductions occur predominantly at low image budgets and moderate-to-high text budgets, consistent with its greater susceptibility to textual perturbations. Kimi exhibits strong reductions across comparatively low allocations in both modalities, whereas InternVL and Nemotron show broader distributions of successful budget combinations. Combined with RQ\textsubscript{1}, this shows that the models differ both in which regime breaks them most easily and in how much perturbation it takes.

Several model--scene configurations show joint perturbations achieving attack success comparable to or higher than the strongest uni-modal regime while distributing the required modification across both modalities, behavior not captured by considering either modality in isolation.

\begin{tcolorbox}[boxrule=0pt,sharp corners,boxsep=2pt,left=2pt,right=2pt,top=2.5pt,bottom=2pt]
\begin{center}
\begin{minipage}[t]{0.99\linewidth}
\textbf{RQ\textsubscript{4} (Modality sensitivity):}
\textit{The evaluated VLMs exhibit distinct modality-sensitivity profiles. Qwen shows the clearest textual susceptibility, Kimi is broadly susceptible across both modalities, while InternVL and Nemotron exhibit more mixed profiles. Joint perturbations further reveal failure-inducing configurations that are not fully characterized by uni-modal evaluation alone, showing that modality sensitivity is best assessed across visual, textual, and joint degradation.}
\end{minipage}
\end{center}
\end{tcolorbox}

\subsection{Threats to Validity}

\head{Construct Validity}
Modality sensitivity is characterized through responses to controlled
perturbations, rather than internal representations. The resulting
profiles describe susceptibility under the evaluated conditions and
do not establish how models internally combine visual and textual
information. Other perturbation families or severity parameterizations
may reveal different patterns.

Visual distance and text-embedding similarity approximate fidelity
but do not guarantee semantic preservation. Image and text budgets
also parameterize different perturbation spaces and are not
metrically equivalent. SWTD combines normalized perturbation measures
with equal weights across active modalities; alternative weighting
schemes may affect comparisons between regimes.

\head{Internal Validity}
The optimization settings, budgets, and search representation influence
which failures are discovered and the modifications required to expose
them. We use the same search algorithm, objectives, parameters, and
budget constraints across models to maintain a consistent protocol.
However, a common configuration may not be equally effective for
every model or perturbation regime.

Baseline filtering produces model-specific input subsets that may
differ in difficulty, limiting direct comparisons across models.
Analyses restricted to outputs containing valid bounding boxes
further exclude empty and malformed responses. The reported
localization, fidelity, and efficiency results should therefore
be interpreted with respect to their eligible subsets.

The human study introduces sampling and annotation uncertainty.
Attention checks filter unreliable responses, but judgments of
interpretability remain subjective, as reflected in the limited
inter-rater agreement. Repeated assessments improve coverage of
individual cases without increasing the number of distinct test
cases evaluated.

\head{External Validity}
The evaluation covers four recent open-source VLMs performing object
localization on three ImageNet-derived scene types and real-world
driving scenes. The observed profiles may not generalize to other
datasets, tasks, proprietary models, or future architectures.
They also depend on the selected prompt template, perturbation
operators, and model configurations. The human study covers a
sample of generated inputs and does not establish interpretability
for the complete generated test set.
\section{Discussion}\label{sec:discussion}
\head{Modality sensitivity as a complementary testing property}
Aggregate robustness scores can conceal differences in the
perturbations required to expose failures. Under our testing
protocol, Qwen exhibits pronounced textual susceptibility,
Kimi produces failures across all regimes, and InternVL and
Nemotron exhibit different degrees of variation across modalities
and scenes. These profiles suggest that models with similar
aggregate robustness may require different testing priorities.
The comparisons characterize behavior under the evaluated
perturbations rather than establish internal reliance on
particular modalities. Establishing whether these behavioral
profiles reflect differences in multimodal integration requires
controlled analysis of internal model behavior.

\head{Architectural interpretation}
The observed differences motivate further investigation of
visual-language integration, training, and output generation.
However, the evaluated models differ along several dimensions,
and the experiments do not isolate their individual effects.
Some architectural relationships nevertheless provide context
for the observed profiles. InternVL3.5-8B initializes its language
model from the Qwen3 series~\cite{wang2025internvl3_5} and shares
the underlying language-model family and text tokenizer with
Qwen3-VL~\cite{Qwen3-VL}, yet the two models exhibit substantially
different textual susceptibility under our protocol.
This makes a simple explanation based solely on language-backbone family or tokenizer less compelling and motivates controlled investigation of multimodal integration and grounding-specific adaptation.

Kimi and Nemotron both use MoE language backbones and exhibit heterogeneous sensitivity profiles. This observation motivates the hypothesis that expert routing or specialization may interact with multimodal perturbations. These correspondences do not establish an architectural cause: the models differ in many other respects, and our small set of architectures does not support an MoE-specific conclusion. Controlled comparisons and analysis of multimodal representations, routing decisions, and expert activations are needed to evaluate this hypothesis.

\head{Implications for testing vision-language models}
In several configurations, joint perturbations achieve failure
rates comparable to or higher than single-modality perturbations
with smaller modifications to each input. Testing should therefore
include image-only, text-only, and joint perturbations, reporting
both failure rates and perturbation magnitude. The resulting
profiles can guide test-suite construction: textual susceptibility
motivates additional prompt-corruption tests, visual susceptibility
motivates broader image-degradation tests, and joint testing
examines combinations that separate evaluations do not cover.
Future work with richer perturbation operators can further examine
whether joint testing exposes distinct failure regions or primarily
reaches failures with smaller per-modality modifications.

Failure categories also require separate reporting. Kimi's high
failure rates frequently reflect malformed structured outputs,
whereas other configurations primarily exhibit localization
degradation or empty responses. Combining these outcomes into
a single score obscures whether remediation should target task
accuracy, generation reliability, or compliance with the output
interface. Distinguishing these failure modes can therefore also
support more targeted debugging and mitigation.

The human study provides complementary evidence of test-case
interpretability. Most assessments judged the sampled inputs
sufficiently clear to perform the task, and mean human localization
accuracy exceeded that of the corresponding VLMs. The observed
failures therefore cannot be attributed solely to inputs becoming
uninterpretable, although limited agreement among assessors
precludes treating all generated cases as valid. Larger-scale
human evaluation could further characterize this boundary.

\head{Implications for VLM evaluation and integration}
Applying a common localization protocol required model-specific
configuration of prompts, output parsing, and bounding-box
coordinates. Gemma-3-4B-IT~\cite{gemma_2025} and
DeepSeek-VL2-Tiny~\cite{wu2024deepseekvl2mixtureofexpertsvisionlanguagemodels}
exhibited high baseline failure rates under this protocol,
limiting their inclusion in subsequent analyses. These outcomes
concern the evaluated configurations rather than the models'
capabilities across tasks.

Differences in coordinate conventions, including pixel and
normalized representations, further complicated comparison.
Together with model-specific prompting and output parsing, these
differences form an adaptation layer that can affect comparative
VLM evaluation. Standardized task interfaces, documented
coordinate conventions, and explicit output-format requirements
would improve reproducibility and reduce the integration effort
required for comparative VLM testing.

\section{Related Work}\label{sec:relw}

We review prior work on uni-modal and multi-modal testing and position our approach with respect to robustness evaluation for VLMs.

\subsection{Testing Methods for Uni-modal DL Systems}

Generalizability testing aims to expose functional failures under realistic or semantically meaningful input variations. Existing approaches include model-based test generation from structured domain abstractions~\cite{Gambi:2019:ATS:3293882.3330566, riccio2020deepjanus, zohdinasab2021deephyperion, 2021-Riccio-ASE} and generative techniques that explore learned input spaces using VAEs, GANs, or diffusion models~\cite{kang2020sinvad, buzhinsky2023metrics, maryam2025benchmarking, merabishvili2026latent}. More recent methods support targeted semantic exploration through feature-space manipulation or generative editing~\cite{weissl2024targeted, chen2026feature, missaoui2023semantic, weissl2026hypernet}. These techniques primarily seek diverse and plausible failure-inducing inputs for uni-modal systems.

Robustness testing instead evaluates sensitivity to perturbations and distribution shifts. Prior work includes metamorphic transformations~\cite{10.1145/3324884.3416584, 10.1145/3238147.3238187}, white-box and black-box input perturbation techniques~\cite{kurakin2018adversarial, croce2020minimally, zhang2020deepsearch}, and generative approaches for producing challenging or realistic adversarial inputs~\cite{RobustifyT2I, baresistoccodiffusion2025}. While these methods provide foundations for systematic robustness evaluation, they target a single input modality and therefore do not characterize how failures depend on the relative contribution of multiple modalities.

\subsection{Testing Methods for Multi-modal DL Systems}

Testing of multi-modal systems extends robustness and generalizability analysis to heterogeneous inputs such as image--text pairs. Recent benchmarks evaluate cross-modal reasoning, compositional understanding, and robustness under distribution shifts or corruptions~\cite{qiu2024benchmarkingrobustnessmultimodalimagetext, sui2025benchmarkingcorruptionrobustnesslvlms}. Generative approaches also synthesize image--text pairs and counterfactual examples to broaden the explored input space~\cite{dillema}. These methods improve test diversity but generally focus on aggregate model performance rather than on how individual modalities contribute to observed failures.

Recent work has further studied VLM robustness under visual and textual perturbations. Jointly corrupting multiple modalities can produce larger performance degradation than perturbing a single modality~\cite{qiu2024benchmarkingrobustnessmultimodalimagetext}; however, existing evaluations commonly rely on predefined corruption levels or evaluate modality-specific perturbations separately~\cite{sui2025benchmarkingcorruptionrobustnesslvlms}. Other studies investigate adversarial attacks in shared embedding spaces~\cite{zhang2025adversarialillusionsmultimodalembeddings, dou2024adversarialattacksmultimodalmodels}, transferable black-box attacks~\cite{hu2025transferableadversarialattacksblackbox, yang2025multifacetedattackexposingcrossmodel}, prompt-injection attacks~\cite{bagdasaryan2023abusingimagessoundsindirect, geng-etal-2025-mathsf}, and attacks targeting inference efficiency or output behavior~\cite{gao2024inducing, wang-etal-2025-vlminferslow}. These approaches primarily quantify whether a model can be made to fail, usually through attack success or task-performance degradation.

Our work addresses a complementary question: \emph{how does failure susceptibility differ across modalities, and what perturbation effort is required to expose it?} We introduce \emph{modality sensitivity testing} and \tool, a behavioral testing framework that systematically compares image-only, text-only, and joint perturbation regimes while continuously searching their severity space. Rather than reducing robustness to a single aggregate score, the resulting modality-sensitivity profile combines attack effectiveness, degradation severity, and modality-specific perturbation requirements, summarized through SWTD. This enables the identification of model-specific visual, textual, and joint vulnerabilities that are not directly characterized by conventional robustness evaluations.
\section{Conclusion}\label{sec:conclusion}

This work presented \tool, a search-based framework for testing vision-language models under visual, textual, and joint input degradations. By systematically exploring these perturbation regimes, \tool generates failure-inducing test cases and supports the characterization of model-specific modality-sensitivity profiles. 

Aggregate robustness scores can obscure substantial differences
in susceptibility to visual, textual, and joint perturbations. Models that solve the same task can differ markedly in which modality exposes failures and in how much perturbation is required to do so. Joint degradation further reveals failure-inducing configurations that are not apparent from considering each modality independently. The human evaluation indicates that many generated inputs remain
interpretable despite inducing VLM failures.

These findings motivate modality-sensitivity testing as a complement to conventional robustness evaluation. Future work includes extending the framework with generative image and text manipulations to explore richer perturbation spaces while retaining the transparency and controllability of the current handcrafted operators. A complementary direction is to investigate the architectural and training factors underlying the observed profiles through controlled white-box studies. In particular, the shared language-model lineage but differing sensitivity of Qwen and InternVL motivates investigation of multimodal integration and grounding-specific adaptation, while the behavior of Kimi and Nemotron motivates analysis of MoE routing and expert specialization. Internal representations, routing decisions, and expert activations could help determine whether these mechanisms contribute to the cross-model differences identified in the study.

\section*{Data Availability}
We provide all code, human survey artifacts, and evaluation scripts in our replication package~\url{https://anonymous.4open.science/r/D265}.

\bibliographystyle{ACM-Reference-Format}
\bibliography{main}
\end{document}